\documentclass[10pt,twocolumn,letterpaper]{article}

\usepackage[pagenumbers]{cvpr} 

\usepackage{booktabs}
\usepackage{times}
\usepackage{amsmath,amssymb,mathtools}
\allowdisplaybreaks
\usepackage{empheq}
\usepackage{graphicx}
\usepackage[font=normalsize]{caption}
\usepackage{subcaption}
\usepackage{floatrow}
\usepackage{enumitem}
\usepackage{xcolor}
\usepackage{multicol}
\usepackage{multirow}
\usepackage{tikz}
\usepackage{graphicx}
\usepackage{array}
\usepackage{colortbl}
\usepackage{tcolorbox}
\usepackage{tabularx}
\usepackage{makecell}
\usepackage{newtxtext,newtxmath}
\usepackage[accsupp]{axessibility}

\usepackage{algorithm}
\usepackage{algpseudocode}

\definecolor{mypast}{RGB}{255, 235, 219}  
\definecolor{mybetween}{RGB}{192, 227, 246}         
\definecolor{myfuture}{RGB}{223, 240, 210}

\usepackage{pifont} 
\definecolor{cvprblue}{rgb}{0.21,0.49,0.74}
\usepackage[pagebackref,breaklinks,colorlinks,allcolors=cvprblue]{hyperref}

\def\paperID{} 
\def\confName{}
\def\confYear{2026}

\title{SeeU: Seeing the Unseen World via 4D Dynamics-aware Generation}

\author{Yu Yuan$^{1}$, Tharindu Wickremasinghe$^{1}$, Zeeshan Nadir$^{2}$, Xijun Wang$^{1}$, Yiheng Chi$^{1}$, Stanley H. Chan$^{1}$ \\
$^1$Purdue University
$^2$Samsung Research America}

\begin{document}

\twocolumn[{%
    \renewcommand\twocolumn[1][]{#1}%
    \maketitle
    \begin{center}
        \captionsetup{type=figure}
        \includegraphics[width=1\textwidth, trim={0 0 0 0}, clip]{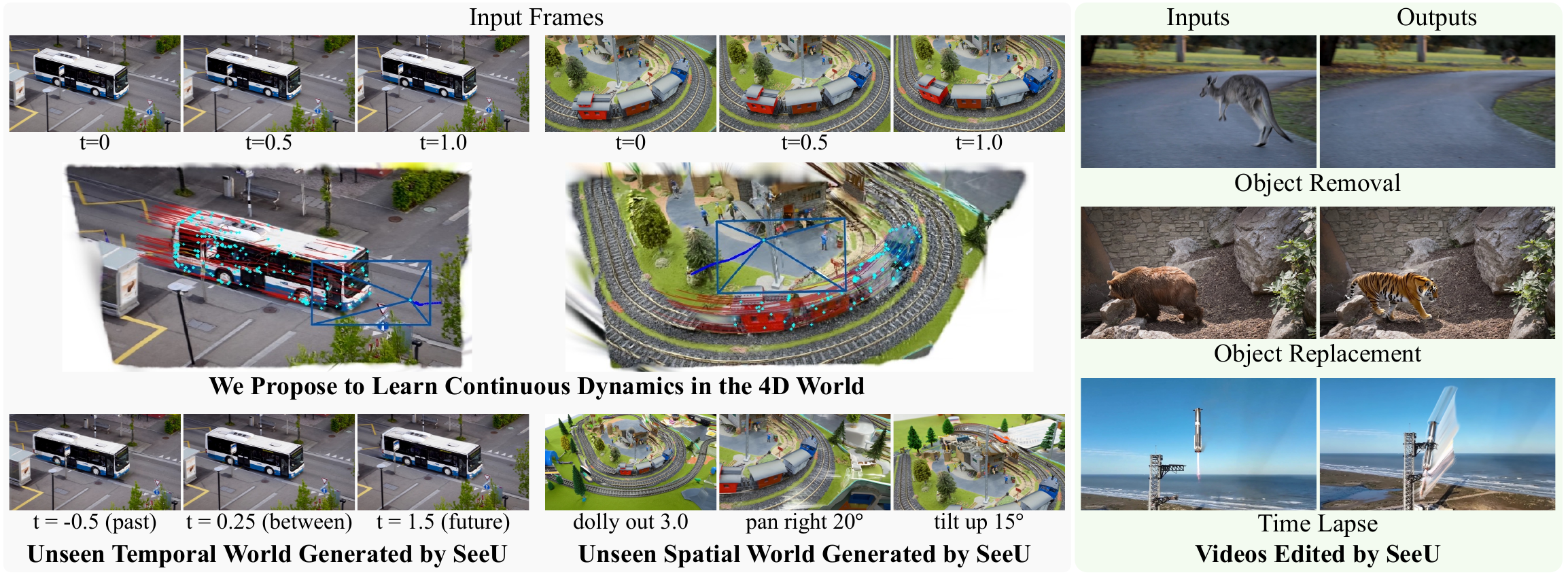}
        \caption{From sparse 2D frames, we learn vanilla and continuous 4D dynamics to better understand scenes and generate unseen worlds across novel times and viewpoints, while enforcing physically plausible motion and consistent 3D geometry.}
        \label{fig:teaser}
    \end{center}

}]

\begin{abstract}
Images and videos are discrete 2D projections of the 4D world (3D space + time). Most visual understanding, prediction, and generation operate directly on 2D observations, leading to suboptimal performance. We propose \textbf{SeeU}, a novel approach that learns the continuous 4D dynamics and generate the unseen visual contents. The principle behind SeeU is a new \textbf{2D$\to$4D$\to$2D} learning framework. SeeU first reconstructs the 4D world from sparse and monocular 2D frames (2D$\to$4D). It then learns the continuous 4D dynamics on a low-rank representation and physical constraints (discrete 4D$\to$continuous 4D). Finally, SeeU rolls the world forward in time, re-projects it back to 2D at sampled times and viewpoints, and generates unseen regions based on spatial-temporal context awareness (4D$\to$2D). By modeling dynamics in 4D, SeeU achieves continuous and physically-consistent novel visual generation, demonstrating strong potentials in multiple tasks including unseen temporal generation, unseen spatial generation, and video editing. All data and code will be public at \href{https://yuyuanspace.com/SeeU/}{here}.
\end{abstract}    
\section{Introduction}
\label{sec:intro}

Humans live in a continuous four-dimensional (4D) space-time and have an exceptional ability to anticipate and imagine environmental evolution. This ability originates from spatial perception \cite{Biederman_1972_Perce}, intuitive physics understanding \cite{Bardes_2024_Revisiting, Garrido_2025_Physics}, and brain mechanism that reconstructs memories to simulate the future \cite{Schacter_2007_Remembering}. A central challenge in machine vision is to endow artificial systems with human-like abilities to perceive, reason about, and generate the dynamic visual world from sensory observations.

To achieve this goal, end-to-end approaches attempt to model dynamics directly from 2D observations. Most methods operate in the pixel domain, learning temporal evolution from large-scale video data, and achieves strong results in video generation \cite{Sora, Blattmann_2023_SVD, Hong_2023_Cogvideo, Yang_2025_Cogvideox, Kong_2024_Hunyuanvideo, Nvidia_2025_Cosmos, wan2025, Peebles_2023_DiT, Wu_2025_Video}, frame interpolation and prediction \cite{Sim_2021_xvfi, Reda_2022_Film, Wang_2025_Inbetween, Feng_2024_inbetweening, Lu_2022_vfiformer, Guo_2024_Gimm, Zhong_2024_Clearer, Zhong_2023_MMVP, Zhong_2024_Mgu} and physics-aware generation \cite{Shi_2024_Motion, Pandey_2025_Motionmode, Niu_2024_Mofa, Xie_2025_Physanimator, Tan_2024_Physmotion, Chefer_2025_VideoJam, Yuan_2025_NewtonGen}. Recent works learn dynamics in low-dimensional latent spaces \cite{Baldassarre_2025_DinoWorld, Karypidis_2025_Dino, Zhou_2024_Dinowm}, which greatly improves computational efficiency. However, these approaches have limitations:
\begin{enumerate}
    \item The observed images and videos are discrete 2D projections of the 4D world onto the camera, directly modeling the dynamics leads to losses of important 3D structures and temporal correlations.  
    \item The observations involve a combination of camera motion and scene dynamics, where the constantly changing camera pose increases the complexity and irregularity of scene motion.
\end{enumerate}
As a result, models trained only on 2D visual patterns without effective 3D or physical supervision often fail to capture the underlying 3D geometry and physical dynamics of the scene. This limitation is more severe in complex out-of-distribution scenarios, such as occlusion, nonrigid deformation, and cluttered 2D projection trajectories.

To address these challenges, we propose to model the continuous 4D dynamics and understand the scenes before visual generation. We present a novel framework, \textbf{SeeU}, that performs visual understanding and generation, embodying this \textbf{2D$\to$4D$\to$2D} paradigm. The motivation behind this innovation is discussed in Section \ref{sec:prelim}. 
SeeU is a departure from existing visual generation methods which focus on dynamics in the projected and discrete 
camera-only space. In SeeU, we first reconstruct a unified 4D representation with tracks from sparse monocular frames to explicitly disentangle camera, static background, and dynamic foreground (2D$\to$4D). Next, continuous 4D dynamics of both the camera and the foreground are learned, producing efficient and physically consistent modeling of scene evolution (discrete 4D$\to$continuous 4D). Finally, with the learned dynamics, we interpolate/extrapolate to unseen times and viewpoints and re-project the 4D representation to 2D to obtain a video skeleton. A spatio-temporal in-context video generator then fills in appearance and details to produce the final video (4D$\to$2D). By enforcing physically consistent 4D representations and dynamics, SeeU enables more interpretable and coherent understanding and generation of dynamic scenes.

We evaluate the proposed framework across a broad spectrum of vision tasks and settings, as illustrated in Fig. \ref{fig:teaser}. In \textbf{unseen temporal} regimes, SeeU enables physically and spatially accurate past frame inference, dynamic frame interpolation, and future frame prediction. In \textbf{unseen spatial} regimes, SeeU synthesizes content for novel camera poses and previously occluded regions, maintaining geometric fidelity and semantic coherence across frames. SeeU also enables \textbf{video editing} applications, including object removal, object replacement, and time-lapse generation, demonstrating versatility and robustness.


In summary, we offer two contributions:
\begin{enumerate}
    \item We introduce a novel concept SeeU, which allows us to see the unseen world through learning the continuous 4D dynamics from the 2D projections. We demonstrate the applicability of SeeU across unseen temporal generation, unseen spatial generation, and video editing.
    \item  The core innovation behind SeeU is a new 2D$\to$4D$\to$2D learning framework that allows us to take 2D inputs and learn the 4D dynamics, and generate the required 2D content. To our knowledge, this flow of information and the corresponding learning scheme is new in the literature.
\end{enumerate}

\section{Why Model Continuous Dynamics in 4D?}
\label{sec:prelim}

As illustrated in Fig.~\ref{fig:prelim_4D}, projecting the continuous 4D world onto 2D images  causes substantial loss of geometric and temporal information. This projection distorts spatial relationships, compresses motion signals, and weakens the physical regularities that govern the observed trajectories. Modeling continuous dynamics directly in the native 4D world has the following advantages:

\begin{figure}[h]
\centering
\includegraphics[width=1.0\linewidth, trim={0 0 0 0}, clip]{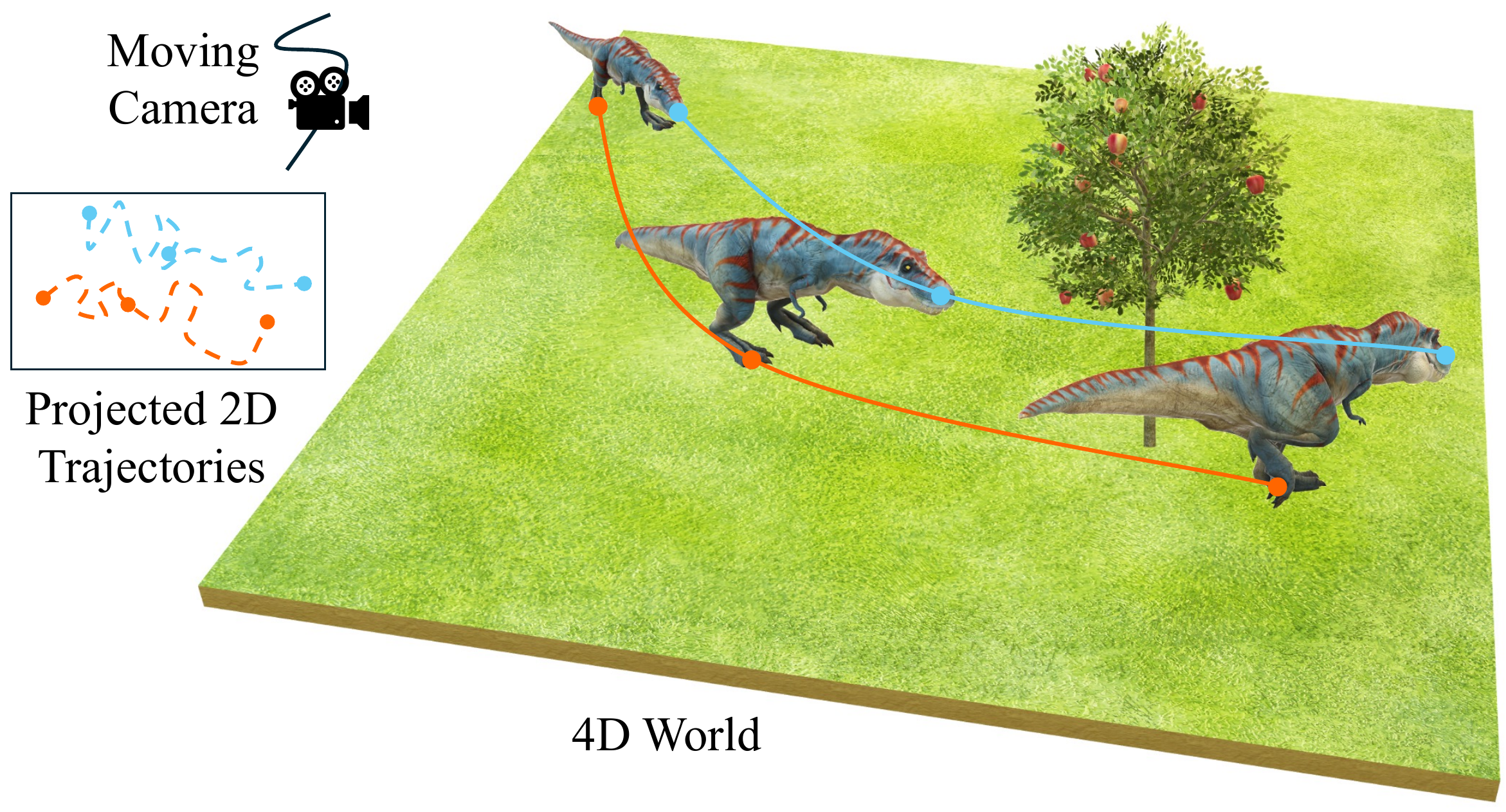}
\caption{Projection and the entanglement of camera and scene motions make recovering accurate 3D geometry and physical trajectories directly from 2D frames particularly challenging; however, these quantities can usually be described in the 4D world explicitly, easily, and elegantly.}
\label{fig:prelim_4D}
\end{figure}

\textbf{(i) 3D Awareness.} Most perception and prediction methods are trained directly on large-scale 2D image/video data. Although they can perform interpolation and future frame prediction to some extent, they struggle in scenarios that require precise 3D reasoning, such as handling geometry, occlusions (see Fig. \ref{fig:unseen_temporal}(b)), and viewpoint changes. The core limitation is the absence of an explicit 3D representation.

\textbf{(ii) Physical Consistency.} The 4D$\to$2D projection removes depth and structural cues and compresses temporal details, making the recovery of true motions from 2D pixel coordinates an ill-posed inverse problem (see the projected 2D trajectories shown in Fig. \ref{fig:prelim_4D}). In practice, many motions are simpler and more structured in 4D, governed by biological or mechanical constraints, classical mechanics, and energy or symmetry principles. These priors can be naturally expressed and learned in 4D. Moreover, casting dynamics as a continuous process in 4D naturally supports interpolation and extrapolation at arbitrary timestamps, leveraging physical priors to produce temporally coherent predictions.

\textbf{(iii) Motion Disentanglement.} When both the camera and the scene are in motion, modeling the dynamics in 2D becomes more difficult because each frame is captured from a different pose where a stable observation coordinate system does not exist. In our framework, the camera, foreground, and background are represented in a unified 4D coordinate system, where their components are explicitly disentangled. Continuous dynamics are then modeled separately for the moving foreground and the camera.

\section{Related Works}
\label{sec:related_works}

\subsection{Dynamic Scene Representation}
\label{sec:related_dynamic_scene}

\textbf{Dynamic NeRFs and 3D Gaussian Splatting.}
Neural radiance field (NeRF) \cite{Mildenhall_2020_Nerf} and 3D Gaussian Splatting (3DGS) \cite{Kerbl_2023_3Dgaussians} are the dominant backbones for 3D reconstruction and novel view synthesis. Recent works extend them to 4D by introducing additional time-dependent dynamics, 
such as a canonical 3D space with a learned deformation field \cite{Li_2020_Neural, Xian_2021_Space, Wang_2021_NeuralTF, Cao_2023_HexPlane, Yang_2024_Gs4d, Wu_2024_4D, Yang_2023_Deformable3dgs, Li_2025_Trace, Zhang_2024_Dynamics, Wu_2025_Difix3d+}. These methods often require a noticeable camera parallax to provide multi-view constraints and tend to struggle when the foreground motion is large \cite{Gao_2022_Dynamic}. Shape-of-Motion \cite{Wang_2025_Som} relaxes these assumptions by injecting depth and 2D tracking priors, improving performance under weaker parallax and faster motion. 

Nonetheless, a common limitation remains for these methods under real-world monocular and temporally sparse inputs: the lack of continuous dynamics modeling leads to poor temporal extrapolation and interpolation (temporally unseen), and causes holes in reconstructions for previously unseen regions (spatially unseen).

\textbf{Dense 3D Point Tracking.}
This emerging class of dynamic scene representation methods tracks dense 2D pixels across frames and back-projects them into 3D using multi-view consistency or depth priors, producing a time-varying 3D point cloud or mesh proxy~\cite{Wang_2023_Omnimotion,Koppula_2024_Tapvid3d,Xiao_2024_SpatialTracker,Xiao_2025_Spatialtracker,Feng_2025_St4rtrack,Ngo_2025_Delta,Ngo_2025_Deltav2,Wang_2024_Scenetracker, Liu_2025_Traceanythingrepresentingvideo, Bian_2025_GSdit}.
Such approaches provide explicit temporal correspondences.

However, their reconstructions are often sparse, noisy, and view-dependent, which limits temporal stability, spatial density, and reliable rendering from novel viewpoints.

\subsection{Dynamics Modeling}

Methods for modeling scene dynamics from videos can be broadly categorized into two types depending on whether the dynamics are learned implicitly or explicitly.

\textbf{Implicit (End-to-End) Dynamics Learning.}
Most existing approaches learn dynamics directly from visual sequences in an end-to-end, data-driven manner. 
They are typically trained on large-scale video datasets using high-capacity sequence models such as Transformers \cite{Vaswani_2017_Transformer, Dosovitskiy_2020_Vit}, multi-layer perceptrons (MLPs) \cite{Tolstikhin_2021_Mixer} or state-space models (e.g., Mamba \cite{Gu_2023_Mamba}).
These methods implicitly capture motion and causal structure either in pixel space by learning frame-by-frame temporal evolution, or in a compact latent space, which improves efficiency.
They have demonstrated strong performance across a wide range of tasks, including world modeling \cite{Baldassarre_2025_DinoWorld, Karypidis_2025_Dino, Zhou_2024_Dinowm}, video generation \cite{Sora, Blattmann_2023_SVD, Hong_2023_Cogvideo, Yang_2025_Cogvideox, Kong_2024_Hunyuanvideo, Nvidia_2025_Cosmos, wan2025, Peebles_2023_DiT, Liu_2024_Exocentric, Wu_2025_Video}, frame interpolation and prediction \cite{Sim_2021_xvfi, Reda_2022_Film, Wang_2025_Inbetween, Feng_2024_inbetweening, Lu_2022_vfiformer, Guo_2024_Gimm, Zhong_2024_Clearer, Zhong_2023_MMVP, Zhong_2024_Mgu}, and physics-aware generation \cite{Shi_2024_Motion, Pandey_2025_Motionmode, Niu_2024_Mofa, Xie_2025_Physanimator, Tan_2024_Physmotion, Chefer_2025_VideoJam, Yuan_2025_NewtonGen}.

\textbf{Explicit (Physics-Informed) Dynamics Learning.}
In contrast, physics-informed approaches incorporate explicit physical constraints into the learning process. 
They embed physical formalisms, such as Hamiltonian or Lagrangian dynamics \cite{Greydanus_2019_Hamiltonian, Lutter_2019_Lagrangian, Zhong_2020_LagrangianDynamics, Deng_2025_Hamiltonian}, into models to infer physical parameters or governing equations directly from videos \cite{Wu_2015_Galileo, Watters_2017_VisualInteraction, Wu_2017_Physics, Belbute_2018_Differentiable, Raissi_2019_Physics, Chari_2019_Physics, Jaques_2020_Physicsvideo, Guen_2020_Disentangling, Hofherr_2023_Video, Garrido_2025_Understanding, Garcia_2025_Video, Li_2025_Nff}.
By enforcing physical consistency, these methods can perform tasks such as future frame prediction and physical parameter estimation, often requiring less data than data-driven approaches.

Despite their success, both categories are fundamentally limited to the 2D domain. After projection to the image plane, much of the true 3D geometry and temporal details is lost, and camera motion becomes entangled with scene dynamics. As a result, these approaches struggle to learn accurate and physically grounded dynamics of the real world.

\subsection{Physics-aware Visual Generation}
Diffusion models \cite{Ho_2020_DDPM, Song_2021_Scorebased} enable high-quality image synthesis \cite{Ramesh_2021_DALLE, Rombach_2022_LDM} and have been extended to video, where large-scale generative models learn motion from Internet-scale datasets \cite{Sora, Blattmann_2023_SVD, Hong_2023_Cogvideo, Yang_2025_Cogvideox, Kong_2024_Hunyuanvideo, Nvidia_2025_Cosmos, wan2025}. Transformer-based diffusion architectures such as DiT \cite{Peebles_2023_DiT} further improve scalability \cite{Kaplan_2020_Scaling} and support powerful systems (e.g., Sora \cite{Sora}). However, simply scaling these models up without effective physical embeddings leads to generated motions that may look plausible but break the real-world dynamics when out-of-distribution \cite{Kang_2025_Farvideogenerationworld, Bansal_2025_Videophy2, Bansal_2024_Videophy, Meng_2025_WorldSimulator, Zhang_2025_Morpheus, Li_2025_Pisa, Gu_2025_Phyworldbench, Chefer_2025_VideoJam, Liu_2025_PhysicalAI, Motamed_2025_PhysicalPrinciples, Lin_2025_Exploringevolutionphysicscognition}.

To improve physical consistency, recent work introduces physical priors into the generation process. They can be classified into three types.
(1) Generation then Physical Simulation: A generator first produces static content and then a physics simulator animates it into videos \cite{Lin_2024_Phys4dgen, Xie_2024_Physgaussian, Tan_2024_Physmotion, Zhang_2024_Physdreamer, Hsu_2024_Autovfx}. This approach is interpretable and controllable, but requires intensive manual setup.
(2) Physical Simulation then Generation: Simulate first to produce trajectories or constraints and then guide a video generator \cite{Yuan_2023_Physdiff, Liu_2024_Physgen, Savantaira_2024_Motioncraft, Chen_2025_Physgen3d, Xie_2025_Physanimator, Li_2025_Wonderplay}. The generator does not learn physics and is dependent on hand-specified rules.
(3) Generation with Learned Physics Priors: physical cues are distilled from pretrained models and used to steer the generator directly \cite{Li_2024_GenerativeImageDynamics, Lv_2024_Gpt4motion, Xu_2024_Motion, Yang_2025_VLIPP, Pandey_2025_Motionmode, Xue_2025_PhyT2V, Cao_2024_Teaching, Wang_2025_Wisa, Yuan_2025_GenPhoto, Zhang_2025_Think, Chefer_2025_VideoJam, Zhang_2025_Videorepa, Feng_2025_Fore, Yuan_2025_NewtonGen}.

Different from above approaches, SeeU first infers deterministic dynamics from multi‐frame observations, forming a physically coherent backbone for video generation.


\section{Proposed Methods}
\label{sec:proposed_methods}
\begin{figure*}[ht]
\centering
\includegraphics[width=0.94\linewidth, trim={0 0 0 0}, clip]{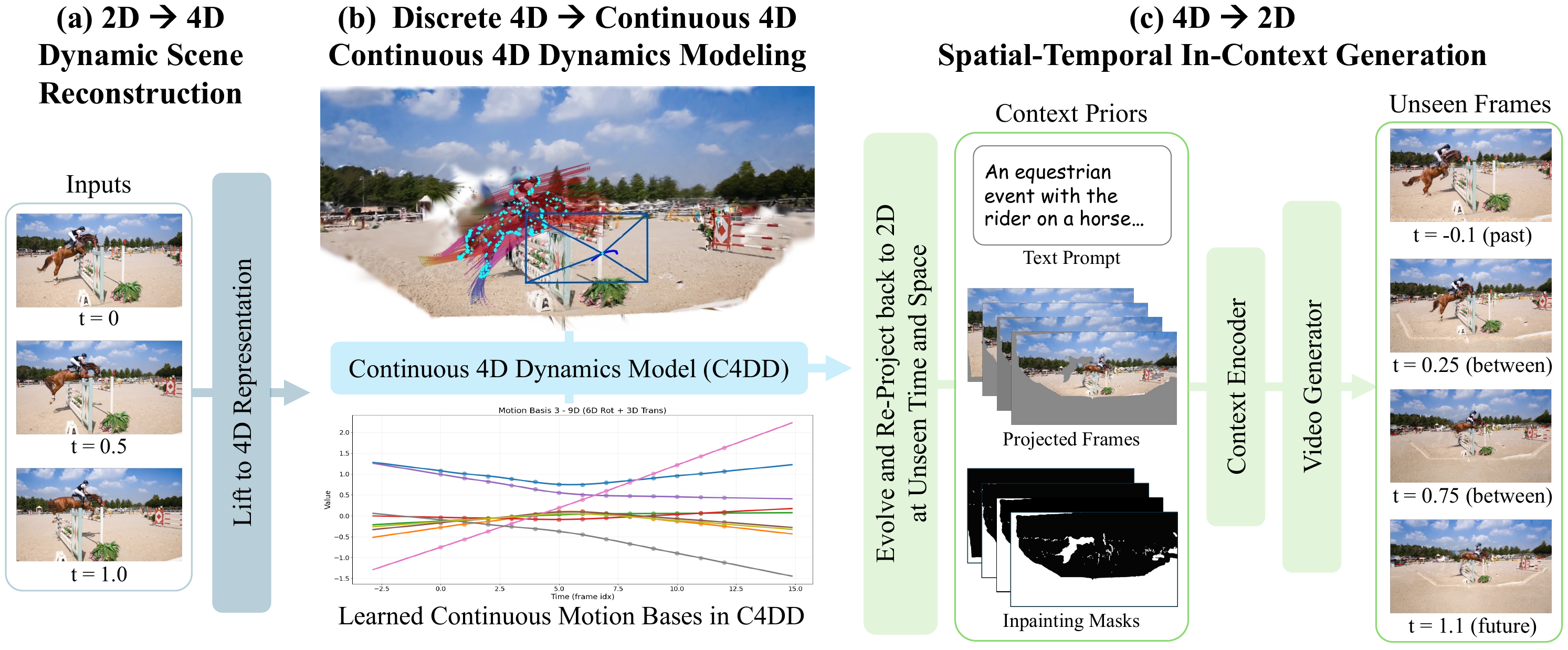}
\caption{\textbf{Pipeline of SeeU.} (a) A dynamic scene is lifted into a 4D representation. (b) Continuous 4D dynamics are learned efficiently with physical and smoothness priors. (c) The learned dynamics evolve the 4D world, which is re-projected to 2D at unseen times and viewpoints; a spatial–temporal in-context video generator completes the unobserved or uncertain areas.}
\label{fig:framework}
\end{figure*}

\subsection{Problem Definition and Overview}

Given a monocular frame sequence $\{I_t \in \mathbb{R}^{H \times W \times 3}\}$ of a dynamic scene, SeeU aims to synthesize novel content at unobserved times and viewpoints beyond the input frames. 

SeeU follows a three-stage pipeline (Fig.~\ref{fig:framework}): 
(i) 2D$\to$4D, (ii) Discrete 4D$\to$Continuous 4D, and (iii) 4D$\to$2D.



\subsection{Dynamic Scene Reconstruction (2D$\to$4D)}
\label{stage1}
In the first stage, SeeU reconstructs the dynamic 4D world with motion trajectories (both dynamic foreground and camera) across time. Our method builds on Shape-of-Motion \cite{Wang_2025_Som} for two specific reasons: 
(1) it is compatible with casual inputs with limited parallax, and (2) explicitly separates static regions from trackable dynamic elements.

We represent the dynamic scene with a set of canonical 3D Gaussians $\{g_0^i\}_{i=1}^N$ that persist over time. 
Each canonical Gaussian is parameterized as
\begin{equation}
g_0^i = (\boldsymbol{\mu}_0^i, \mathbf{R}_0^i, \mathbf{s}^i, o^i, \mathbf{c}^i),
\end{equation}
where $\boldsymbol{\mu}_0^i \in \mathbb{R}^3$ denotes the canonical mean, 
$\mathbf{R}_0^i \in \mathbb{SO}(3)$ the canonical orientation, 
$\mathbf{s}^i \in \mathbb{R}^3$ the scale, 
$o^i \in \mathbb{R}$ the opacity, 
and $\mathbf{c}^i \in \mathbb{R}^3$ the color.
To model changes, each Gaussian evolves from the canonical frame $t_0$ to frame $t$ via a per-frame rigid transformation 
$\mathbf{T}_{0 \to t} = [\mathbf{R}_{0 \to t}, \mathbf{t}_{0 \to t}] \in  \mathbb{SE}(3)$:
\begin{equation}
\boldsymbol{\mu}_t^i = \mathbf{R}_{0 \to t}\boldsymbol{\mu}_0^i + \mathbf{t}_{0 \to t}, 
\quad 
\mathbf{R}_t^i = \mathbf{R}_{0 \to t}\mathbf{R}_0^i.
\end{equation}

Specifically, for the input frames $I_t$, 
we first estimate the camera intrinsics, extrinsics, and per-frame depth using MegaSaM~\cite{Li_2025_MegaSaM}. 
We then obtain segmentation masks of moving foreground objects with Track-Anything~\cite{Yang_2023_Track}, 
and extract 2D point tracks using TAPIR~\cite{Doersch_2023_Tapir}. 
Finally, we fuse the RGB frames with these 2D priors to infer a dynamic scene representation, 
and output the frame-level camera pose and attributes of each foreground Gaussian properties $\mathbf{P}$ 
(\textit{e.g.}, position, orientation, and scale).

\subsection{Continuous 4D Dynamics Modeling (Discrete 4D$\to$Continuous 4D)}

With the frame-level camera poses and the per-frame attributes of each foreground Gaussian obtained from Stage~1, our goal is to recover physically credible, continuous-time 4D dynamics from these discrete observations.
Let $\mathbf{P}_t^{i}$ denote the attributes of the $i$-th Gaussian at frame $t$,
and let $\mathbf{C}_t$ be the camera pose at frame $t$. We seek continuous trajectories
 for the Gaussians and the camera.

This task presents two main challenges: (1) \textbf{Efficiency.} The number of foreground Gaussians can be large (e.g., $80\mathrm{k}$), 
    making it infeasible to learn an independent time-varying trajectory for each primitive.
(2) \textbf{Physical Consistency.} The recovered trajectories should be smooth and physically plausible, avoiding abrupt, nonphysical jumps in position, rotation, velocity, or acceleration.


To address challenge (1), we employ a low-rank motion parameterization inspired by Shape-of-Motion \cite{Wang_2025_Som}.
For any foreground Gaussian $i$ at time $t$, its properties $\mathbf{P}_t^{i}$ are
\begin{equation}
\mathbf{P}_t^{i} = \mathbf{P}_0^{i} + \underbrace{\mathbf{B}(t)}_{\in\,\mathbb{R}^{m\times K}} \underbrace{\mathbf{w}_i}_{\in\,\mathbb{R}^{K}},
\quad
\mathbf{P}_0^{i},\, \mathbf{P}_t^{i} \in \mathbb{R}^{m}.
\end{equation}
where $\mathbf{P}_0^{i}$ is the initial state of Gaussian $i$, $\mathbf{B}(t)$ is a set of global motion basis functions shared by all Gaussians, and $\mathbf{w}_i$ is a time-invariant coefficient vector for Gaussian $i$. 
Through this parameterization, instead of directly learning $N$ separate trajectories, we learn only a small number $K$ of shared basis functions $\mathbf{B}(t)$ (where $K \ll N$), and each Gaussian is represented by few coefficients. The discrete motion bases are first initialized from 3D point trajectories using Procrustes analysis, and subsequently refined by minimizing the photometric reconstruction error.

To address challenge (2), we observe that in natural sequences without sudden disturbances, both the per motion basis trajectory and the camera pose trajectory points in $\mathbb{SE}(3)$, exhibit simple and smooth temporal trends shown in Fig.~\ref{fig:framework}(b), despite complex motions in raw videos. This motivates modeling them as continuous time functions parameterized by B-spline curves.


Motivated by the above analysis, we formulate a lightweight, physics-aware 
\textbf{Continuous 4D Dynamics Model (C4DD)} that explicitly learns the control points 
$\{\mathbf{q}_j\}_{j=1}^M$ of the B-spline functions
\begin{equation}
\label{eq:Bt-sp}
\hat{\mathbf{B}}_{t}
\;=\;
\sum_{j=1}^{M} N_{j,d}(t)\,\mathbf{q}_j ,
\end{equation}
where $\{N_{j,d}(t)\}$ are the B\mbox{-}spline basis functions and 
$\mathbf{q}_j$ are the learnable control points for the motion bases. 
The number of control points $M$ controls the curve capacity: larger $M$ allows richer temporal variations, while smaller $M$ enforces stronger smoothness and regularization. 

During training, both the camera and the shared 
motion bases are optimized jointly under a hybrid objective:
\begin{equation}
\label{eq:ltotal}
\mathcal{L}_{\text{total}}
\;=\;
\underbrace{\mathcal{L}_{\text{data}}}_{\text{data loss}}
\;+\;
\lambda_{\text{phys}}\,
\underbrace{\mathcal{L}_{\text{phys}}}_{\text{physical loss}} .
\end{equation}

The data term enforces consistency between the C4DD-estimated motion bases and the
discrete observations obtained in  challenge (1):
\begin{equation}
\label{eq:ldata}
\mathcal{L}_{\text{data}}
=
\sum_{t \in \mathcal{T}_{\text{obs}}}
\big\|
\hat{\mathbf{B}}_{t} - \mathbf{B}_{t}^{\text{obs}}
\big\|_2^2 ,
\end{equation}
where $\mathbf{B}_{t}^{\text{obs}}$ denotes the observed (discrete) motion bases at
timestamp $t$, and $\hat{\mathbf{B}}_{t}$ is the C4DD prediction evaluated from the
B\mbox{-}spline parameterization:

To further enforce physical plausibility, we regularize the temporal derivatives 
of both the motion bases and the camera trajectories. 
The physical loss penalizes rapid changes in translation and rotation, 
with larger weights assigned to extrapolated time intervals:
\begin{equation}
\label{eq:lphys}
\begin{aligned}
\mathcal{L}_{\text{phys}}
= \mathbb{E}_{\tau_{\text{ex}}(t)} \Big[
&\|\ddot{\mathrm{MB}}_{\text{trans}}(t)\|_2^2
+\|\ddot{\mathrm{Cam}}_{\text{trans}}(t)\|_2^2 \\[3pt]
&+\mathbb{I}_{\text{rot}}\;\|\ddot{\mathrm{Cam}}_{\text{rot}}(t)\|_2^2
\Big],
\end{aligned}
\end{equation}
where $\ddot{\mathrm{MB}}_{\text{trans}}(t)$ and $\ddot{\mathrm{Cam}}_{\text{trans}}(t)$ 
denote the second-order temporal derivatives of the translational components for the motion bases and the camera trajectory, respectively; 
$\ddot{\mathrm{Cam}}_{\text{rot}}(t)$ is the rotational acceleration; 
$\mathbb{I}_{\text{rot}}\!\in\!\{0,1\}$ toggles rotation regularization; 
$\tau_{\text{ex}}(t)$ assigns higher weights to unobserved (extrapolated) time spans 
in the past or future. 

\subsection{Spatial-Temporal In-Context Generation (4D$\to$2D)}

Leveraging the continuous 4D dynamics learned in Stage 2 together with the Stage 1 scene representation, we evolve the scene to any timestamp. We then either position the camera at a desired pose or evolve its pose via C4DD and render the resulting 2D projection, which acts as a video scaffold encoding precise geometry, motion, and occlusion structure.

Note that the scaffold may be incomplete, as certain regions can be missing or unreliable, including: (1) areas never observed (novel viewpoints or previously occluded regions), (2) locations where the projected Gaussians yield low confidence, and (3) thin structures and sharp depth discontinuities that can introduce projection artifacts (e.g., along object boundaries and occlusion edges).


To address these issues, we leverage the strong spatial–temporal in-context capabilities \cite{Wiedemer_2025_Video, Bian_2025_VideoPainter, Jiang_2025_VACE} of video generation models to repair frames (Fig. \ref{fig:framework}(c)). 
We supply the model with three types of context priors:
(1) a structured prompt derived from a vision–language model (VLM) \citep{Wang_2024_Tarsier} as a caption of the scene, which encodes global semantics, and specifies the inpainting content, (2) projected frames that serve as a geometric and photometric reference, and (3) per-frame inpainting masks detected during projection that mark uncertain or missing regions. A Context Encoder ingests these priors, produces context embeddings, and injects them into a pre-trained video generator.
With these context priors, the video generator fills in high-frequency details, textures, and temporal coherent contents in regions uncertain or unseen in the original observations.

\section{Experiments}
\begin{figure*}[t]
\centering
\includegraphics[width=0.97\linewidth, trim={0 0 0 0}, clip]{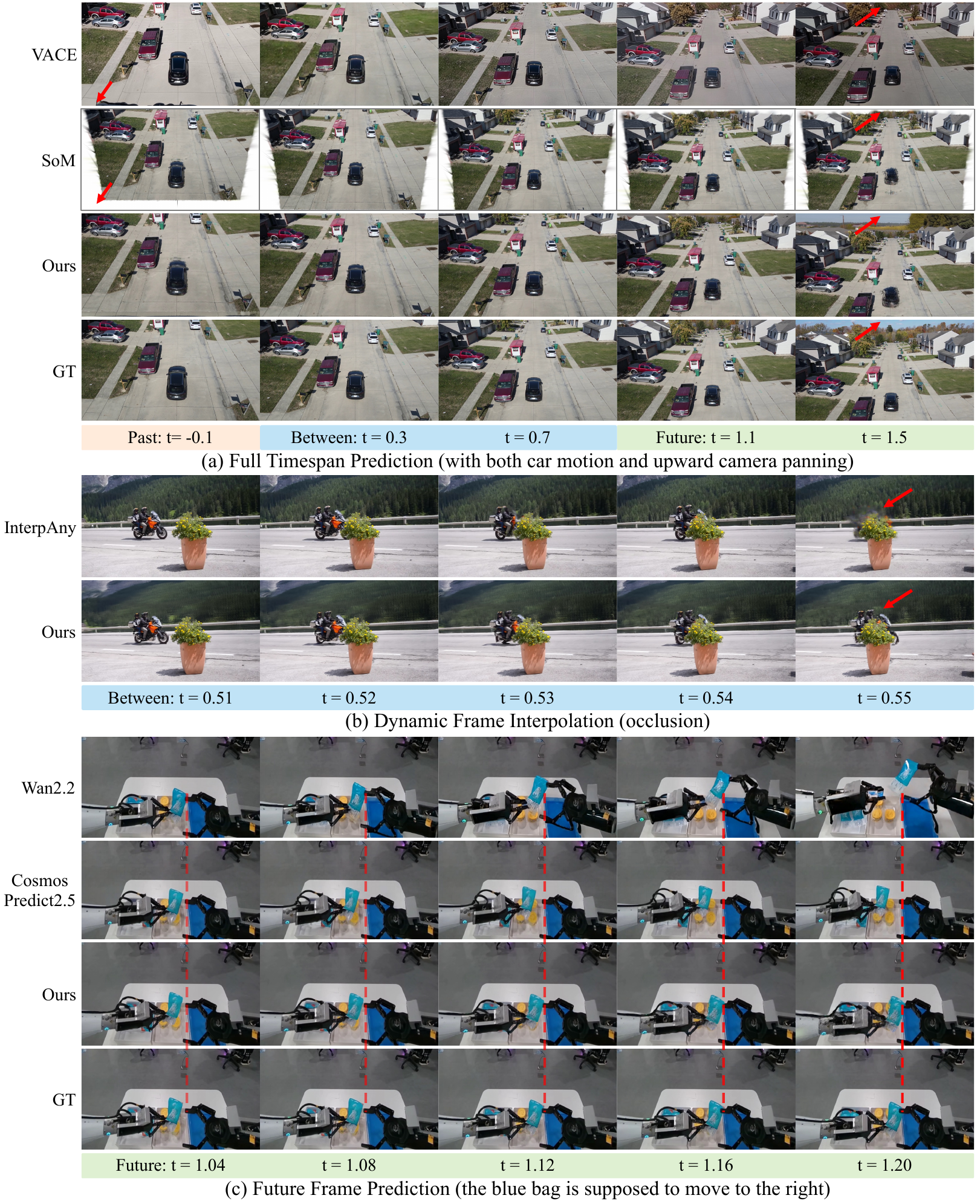}

\caption{\textbf{Visual comparisons on unseen temporal generation.} SeeU supports continuous-time generation across the entire time span (\colorbox{mypast}{past}–\colorbox{mybetween}{between}–\colorbox{myfuture}{future}), yielding more physically plausible motion and stronger geometric consistency.}
\label{fig:unseen_temporal}
\end{figure*}

\begin{figure}[ht]
\centering
\includegraphics[width=1.0\linewidth, trim={0 0 0 0}, clip]{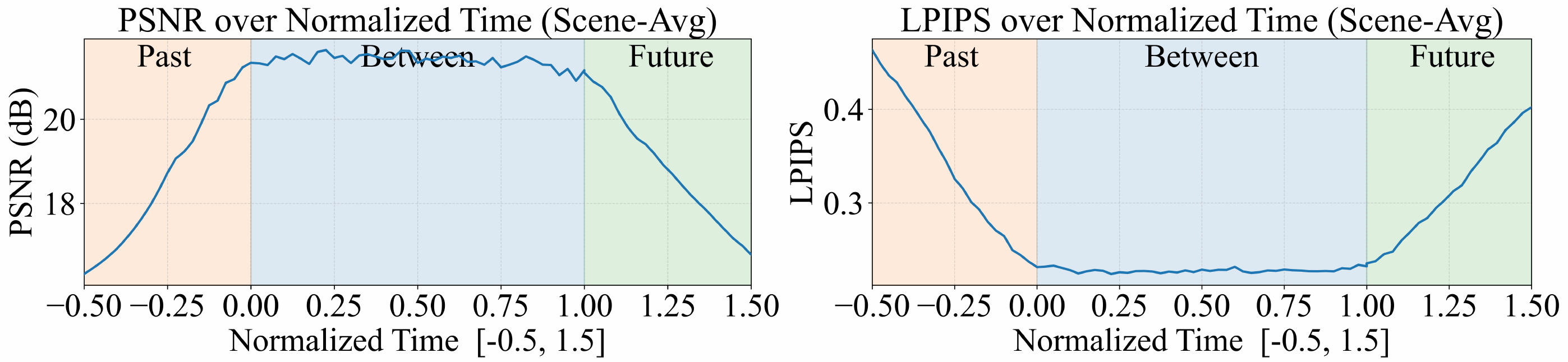}
\caption{\textbf{Temporal prediction error analysis.}}
\label{fig:time_error}
\end{figure}

\begin{figure*}[ht]
\centering
\includegraphics[width=0.97\linewidth, trim={0 0 0 0}, clip]{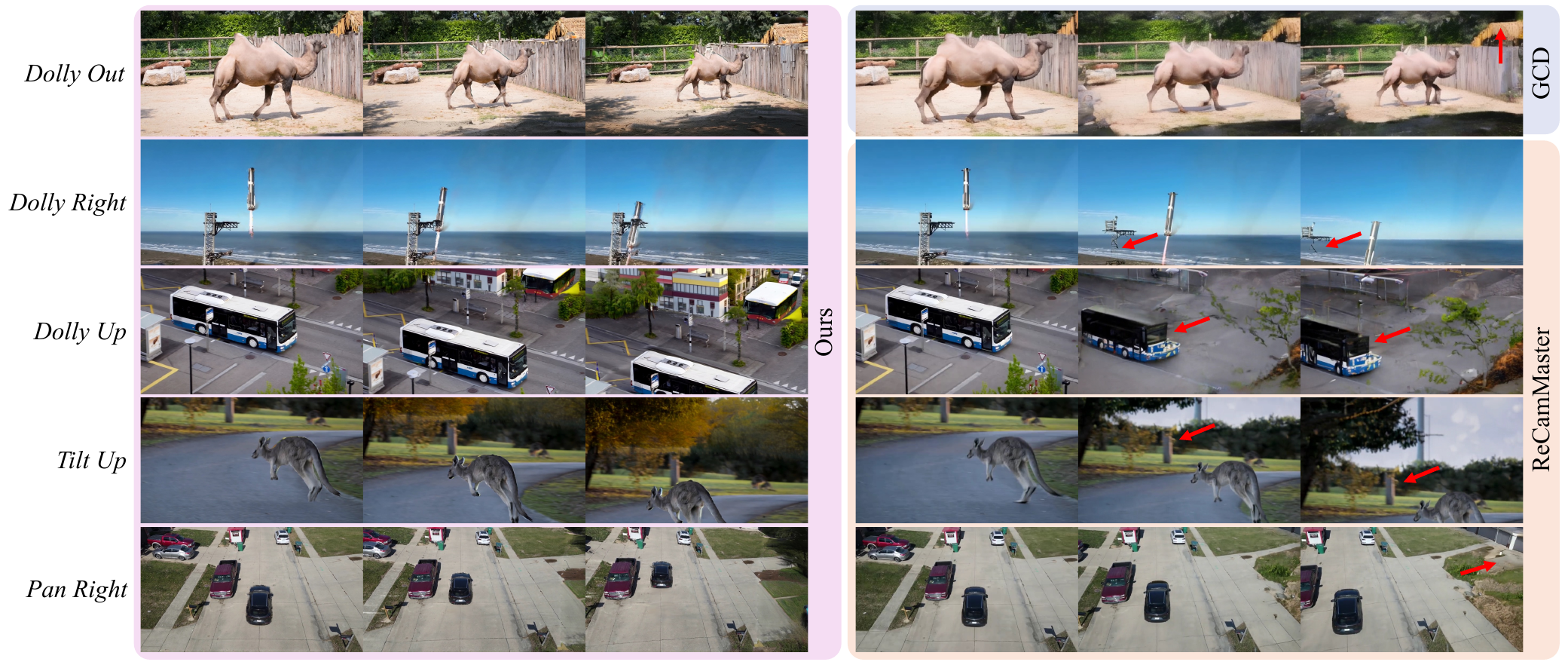}
\caption{\textbf{Visual comparisons on unseen spatial generation.} SeeU exhibits strong 3D awareness and scene consistency.}
\label{fig:unseen_spatial}
\end{figure*}

\begin{table*}[ht]
    \centering
    \resizebox{\textwidth}{!}{
    \small
    \begin{tabular}{lcccccccccccc}
        \Xhline{3\arrayrulewidth}
      \multirow{2}{*}{Methods} 
           & \multicolumn{4}{c}{Past (6.67\%) Frame Inference} 
           & \multicolumn{4}{c}{Dynamic Frame Interpolation} 
           & \multicolumn{4}{c}{Future (6.67\%) Frame Prediction}  \\
        \cmidrule(lr){2-5}\cmidrule(lr){6-9}\cmidrule(lr){10-13}
        & PSNR $\uparrow$ & SSIM $\uparrow$ & LPIPS $\downarrow$ & C-LPIPS 
        & PSNR $\uparrow$ & SSIM $\uparrow$ & LPIPS $\downarrow$ & C-LPIPS 
        & PSNR $\uparrow$ & SSIM $\uparrow$ & LPIPS $\downarrow$ & C-LPIPS  \\
        \hline

      Reference  
            & -- & -- & -- & 0.0397  
            & -- & -- & -- & 0.0356
            & -- & -- & -- & 0.0389 \\

     \hline 
       InterpAny \cite{Zhong_2024_Clearer}  
            & -- & -- & -- & -- 
            & 20.5406 & 0.5636 & 0.2420 & 0.0525
            & -- & -- & -- & -- \\

       Wan 2.2 \cite{wan2025}  
            & -- & -- & -- & -- 
            & -- & -- & -- & -- 
            & 18.2989 & 0.5064 & 0.2559 & 0.0704\\

       Cosmos \cite{Nvidia_2025_Cosmos} 
            & -- & -- & -- & --
            & -- & -- & -- & --
            & 20.0672 & 0.5557 & 0.2885 & 0.0355\\

       SoM \cite{Wang_2025_Som} 
            & 15.5464 & 0.4518 & 0.3876 & 0.0116
            & 16.3650 & 0.4814 & 0.3555 & \textbf{0.0340}
            & 15.4301 & 0.4585 & 0.3885 & 0.0141 \\

       VACE \cite{Jiang_2025_VACE} 
            & 17.1430 & 0.4648 & 0.3673 & 0.0924
            & 18.1617 & 0.5041 & 0.3587 & 0.1821
            & 17.7080 & 0.4938 & 0.3543 & 0.1244 \\

       Ours 
            & \textbf{20.4682} & \textbf{0.5496} & \textbf{0.2484} & \textbf{0.0355}
            & \textbf{21.0726} & \textbf{0.5715} & \textbf{0.2271} & 0.0327
            & \textbf{20.5353} & \textbf{0.5581} & \textbf{0.2431} & \textbf{0.0358} \\
        \Xhline{3\arrayrulewidth}
    \end{tabular}}
    \caption{
    \textbf{Quantitative comparison on unseen temporal generation.} Metrics are averaged across all scenes. The extrapolation windows (past and future) each covers 6.67\% of the sequence duration. Best values are shown in \textbf{bold}. Note that C-LPIPS is not lower-better; it should be close to the reference value.
    }
    \label{tab:unseen_temporal}
\end{table*}

\begin{table*}[ht]
    \centering
    \resizebox{\textwidth}{!}{
    \small
    \begin{tabular}{lccccccccccccccc}
        \Xhline{3\arrayrulewidth}
        \multirow{2}{*}{Methods} 
            & \multicolumn{3}{c}{Dolly Out} 
            & \multicolumn{3}{c}{Dolly Right} 
            & \multicolumn{3}{c}{Dolly Up} 
            & \multicolumn{3}{c}{Tilt Up} 
            & \multicolumn{3}{c}{Pan Right}  \\
        \cmidrule(lr){2-4}\cmidrule(lr){5-7}\cmidrule(lr){8-10}\cmidrule(lr){11-13}\cmidrule(lr){14-16}
        
        & EE $\downarrow$ & EIR $\uparrow$ & CLIP-V $\uparrow$
        & EE $\downarrow$ & EIR $\uparrow$ & CLIP-V $\uparrow$
        & EE $\downarrow$ & EIR $\uparrow$ & CLIP-V $\uparrow$
        & EE $\downarrow$ & EIR $\uparrow$ & CLIP-V $\uparrow$
        & EE $\downarrow$ & EIR $\uparrow$ & CLIP-V $\uparrow$ \\
        \hline

        GCD \cite{Vanhoorick_2024_GCD} 
            & 0.2004 & 0.5129 & 0.9253
            & -- & -- & --
            & -- & -- & --
            & -- & -- & --
            & -- & -- & -- \\

        ReCamMaster \cite{Bai_2025_Recammaster}
            & 0.2382 & 0.6736 & 0.9366
            & 0.2130 & 0.7049 & 0.9227
            & 0.2087 & 0.7053 & 0.9248
            & 0.2281 & 0.7161 & 0.9256
            & 0.2146 & 0.7035 & 0.9250 \\

        Ours
            & \textbf{0.1997} & \textbf{0.7848} & \textbf{0.9690}
            & \textbf{0.1895} & \textbf{0.7947} & \textbf{0.9542}
            & \textbf{0.1908} & \textbf{0.7841} & \textbf{0.9695}
            & \textbf{0.2082} & \textbf{0.8208} & \textbf{0.9466}
            & \textbf{0.1891} & \textbf{0.8229} & \textbf{0.9574} \\
        \Xhline{3\arrayrulewidth}
    \end{tabular}}
    \caption{
    \textbf{Quantitative comparison on unseen spatial generation.}
    Lower Epipolar Error (EE) and higher Epipolar Inlier Ratio (EIR) indicate better 3D geometric consistency, and higher CLIP-V refers to higher scene consistency.}
    \label{tab:unseen_spatial}
\end{table*}

\begin{table}[ht]
    \resizebox{\linewidth}{!}{
    \begin{tabular}{lccccc}
        \Xhline{3\arrayrulewidth}
      \multirow{3}{*}{Methods} 
           & \multicolumn{3}{c}{Unseen Temporal} & \multicolumn{2}{c}{Unseen Spatial}  \\

           \cmidrule(lr){2-4}\cmidrule(lr){5-6}

        & PSNR $\uparrow$ & SSIM $\uparrow$ & LPIPS $\downarrow$
        & EE $\downarrow$
        &  CLIP-V $\uparrow$ \\

        \hline
        C4DD w/ MLP & 17.5372 & 0.3934 & 0.4274 & 0.3125 & 0.7390 \\
        
        w/o physics loss & 19.3593 & 0.5274 & 0.2735 &  0.2240 & 0.9198 \\

        \hline
        
       Ours (5 frames)  & 18.3583 & 0.4521 & 0.3054 & 0.2852 & 0.9284 \\

       Ours (10 frames) & 20.1626  & 0.5378 & 0.2507  & 0.2044 & 0.9551
       \\
        
       Ours (15 frames) & 20.3870 & 0.5484 & 0.2408 & 0.2003 & 0.9581 \\
       
       Ours (20 frames) & \textbf{21.0832} & \textbf{0.5522} & \textbf{0.2387} & \textbf{0.1972} & \textbf{0.9596} \\      

        \Xhline{3\arrayrulewidth}
    \end{tabular}}
    \caption{\textbf{Quantitative results of ablation study.}}
    \label{tab:ablation}
\end{table}

We evaluate SeeU in novel content generation tasks where major challenges originate from \emph{unseen time} (e.g., past, between-frames, and future) and \emph{unseen space} (e.g., camera motion and occlusion). We also demonstrate its broader potential in video editing. Section \ref{sec:implementation} presents implementation details, Section \ref{sec:comparisons} compares SeeU with other baselines, Section \ref{sec:ablation} discusses ablation results, and Section \ref{sec:more_apps} focuses on video editing applications.

\subsection{Implementation Details}
\label{sec:implementation}

\textbf{Datasets.}
We evaluate our framework on \textsc{SeeU45}, a curated set of 45 dynamic scenes assembled from our own captures and public sources, including video tracking dataset TAP-vid \cite{Doersch_2022_Tapvid, Perazzi_2016_Davis}, high frame-rate I2-2000FPS  \cite{Chennuri_2024_Quanta}, robotics video set AgiBot World \cite{Bu_2025_Agibot}, and Animal Kingdom dataset \cite{Ng_2022_Animal}. 
The collection of scenes spans indoor/outdoor environments, diverse subjects (animals, robots, humans, everyday objects), and a wide range of camera regimes (static, handheld) and motion types (rigid/non-rigid).

\textbf{Training Details.} 
SeeU is trained and run in three stages sequentially. All stages are trained with Adam optimizer on a single NVIDIA A100 (80\,GB) GPU. In the first stage (2D$\to$4D), the numbers of Gaussian primitives are $80{,}000$ for the dynamic foreground and $80{,}000$ for the background, and 10 motion bases are learned. The learnable components are optimized for $4{,}000$ iterations. For a typical $10$-frame sequence of $960\times540$ resolution, this optimization takes approximately $1$ hour. In the second stage (Discrete 4D$\to$Continuous 4D), we train the Continuous 4D Dynamics Model (C4DD) using cubic ($\deg=3$) B-splines with $8$ control points and a physics loss weight of $\lambda_{\text{phy}}=1\times10^{-4}$. The learning rate is $1\times10^{-5}$, and the batch size is $64$. Training $1{,}000$ epochs takes $\sim\!10$ minutes. In the third stage (4D$\to$2D), we perform a computationally efficient fine-tuning of VACE \cite{Jiang_2025_VACE} on our multi-semantic-mask distribution, which takes approximately $2$ hours. 


\textbf{Metrics.} For unseen time, we evaluate physical accuracy and frame-to-frame consistency, and, for unseen space, 3D geometry and scene-level coherence. To evaluate temporally unseen generation, we sample a subset of intermediate frames per scene as input and generate their (i) past, (ii) in-between, and (iii) future frames. We compare the generation results with their temporally nearest frames in the full sequence, reporting \textsc{PSNR}, \textsc{SSIM}, \textsc{LPIPS} \cite{Zhang_2018_Perceptual}, and cross-frame consistency C-LPIPS \cite{Yuan_2025_GenPhoto}. To evaluate spatially unseen generation, we measure (i) 3D geometric agreement using Epipolar Error (EE) and Epipolar Inlier Ratio (EIR) (details in Suppl. Materials) and (ii) semantic consistency between source and target frames using CLIP-V \cite{Kuang_2024_CVD}.

\subsection{Results}
\label{sec:comparisons}

\textbf{Temporally Unseen Generation.} We evaluate performance on three sub-tasks: past frame inference, dynamics frame interpolation (between-frames), and future frame prediction. As currently no single baseline can comprehensively address all three tasks, we benchmark SeeU against a suite of specialized state-of-the-art (SOTA) methods, including video frame interpolation model InterpAny \cite{Zhong_2024_Clearer}, future predictor COSMOS Predict2.5 \cite{Nvidia_2025_Cosmos}, and image-to-video generator Wan2.2 \cite{wan2025} conditioned on the last frame. In addition, we extend VACE \cite{Jiang_2025_VACE} to cover the full temporal range by masking unseen frames, and Shape-of-Motion \cite{Wang_2025_Som} by performing linear interpolation and extrapolation. As shown in Fig. \ref{fig:unseen_temporal} and Table \ref{tab:unseen_temporal}, SeeU demonstrates stronger continuous-time dynamics awareness and 3D geometric interpretation, outperforming baselines in temporal extrapolation and dynamic interpolation with more physically plausible motion and higher geometric consistency. Fig. \ref{fig:time_error} presents the temporal prediction error analysis over the time span from the past 50\% to the future 50\%. We observe a decrease of accuracy in the extrapolated regions approximately linear to the temporal distance.

\textbf{Spatially Unseen Generation.} 
We compare SeeU with specialized camera-controllable video models GCD \cite{Vanhoorick_2024_GCD} and ReCamMaster \cite{Bai_2025_Recammaster}. The evaluation is performed by experimenting on five canonical camera motions: \emph{dolly-out}, \emph{dolly-right}, \emph{dolly-up}, \emph{tilt-up}, and \emph{pan-right}. As illustrated in Fig. \ref{fig:unseen_spatial}, SeeU produces temporally smoother results, providing more coherent visual continuity and generating more accurate spatial occlusions. Table \ref{tab:unseen_spatial} demonstrates that SeeU achieves higher geometric fidelity (EE and EIR) and scene consistency (CLIP-V) than all baselines, reflecting its capability to render richer scene details.

\subsection{Ablation Study}
\label{sec:ablation}

\textbf{C4DD Architecture.}
We replace the B\hbox{-}spline parameterization with plain MLP layers. 
Tabel \ref{tab:ablation} indicates that this variant produces results with reduced temporal smoothness and physical consistency, confirming the advantageous inductive bias of spline priors for continuous dynamics.

\textbf{Physics-Informed Loss.}
Table \ref{tab:ablation} shows that setting the physics loss weight $\lambda_{\text{phy}}$ to $0$ degrades frame consistency, highlighting the role of physics constraints in stabilizing continuous dynamics.

\textbf{Input Sparsity.}
As shown in Table. \ref{tab:ablation}, we investigate the effects of sparse observations by subsampling the input to $5/10/15/20$ frames.
C4DD maintains stable performance as frame density decreases, demonstrating robustness to sparse inputs and preservation of temporal continuity.

\subsection{Application to Video Editing}
\label{sec:more_apps}

By constructing a continuous, unified model of time and 3D space, SeeU enables a variety of video editing applications, as shown in Fig. \ref{fig:teaser}. For example, \textbf{object removal} can be performed by deleting foreground Gaussians and inpainting the revealed background regions by de-occlusion. By replacing the original foreground content and conditioning on a new text prompt, SeeU also enables controllable \textbf{object replacement} in videos. Leveraging the learned continuous 4D dynamics, we can further re-parameterize foreground trajectories in time to create \textbf{time-lapse} effects.


\section{Conclusions}
SeeU implements a 2D$\to$4D$\to$2D pathway to achieve a unified understanding and generation. The technical core of SeeU is the construction of physically-consistent, continuous 4D dynamics within the native 4D world to describe its dynamic evolution. Combined with in-context video generation, SeeU synthesizes previously unseen content across time and space, enabling advances in world modeling and narrowing the gap between physical and generative domains.\\
\textbf{Limitation:} SeeU is applicable to a broad range of real-world dynamic scenes; however, given the limitations of the underlying modules (tracking, camera estimation, and 4D reconstruction), this paper focuses on inputs with pronounced, smooth, and temporally stable foreground motion.

\subsubsection*{Acknowledgments}
This work is supported, in part, by the United States National Science Foundation under the grants 2133032, 2431505, and a research award from Samsung Research America. 

We thank Yichen Sheng and Lu Ling for helpful discussion.


{
    \small
    \bibliographystyle{ieeenat_fullname}
    \bibliography{main}
}

\clearpage
\setcounter{page}{1}
\maketitlesupplementary

\section{Introduction}

This supplementary material provides additional discussions and details on the SeeU45 data (Section \ref{sec:supp_data}), Continuous 4D Dynamics Model (C4DD) design and ablation study (Section \ref{sec:supp_C4DD}), spatial-temporal in-context generation (Section \ref{sec:supp_context}), the 3D geometric metrics used in the experiments (Section \ref{sec:supp_metrics}), the limitations (Section \ref{sec:supp_limitations}), robustness and scalability analysis (Section \ref{sec:supp_complex}), more comparison results (Section \ref{sec:supp_comparison}), and more visual results (Section \ref{sec:supp_results}).

To more clearly demonstrate SeeU’s temporal and spatial generation abilities, we recommend that readers refer to the \textbf{videos} included in the Project Page at \href{https://yuyuanspace.com/SeeU/}{https://yuyuanspace.com/SeeU/}.


\section{More Details of SeeU45 Data}
\label{sec:supp_data}
The SeeU45 dataset consists of 45 dynamic scenes, including 10 scenes that we manually captured and 35 scenes collected from public video datasets \cite{Doersch_2022_Tapvid, Perazzi_2016_Davis, Chennuri_2024_Quanta, Bu_2025_Agibot, Ng_2022_Animal}. Each scene is provided in two forms: a ground-truth (GT) sequence and a training subset. The GT split contains the full dynamic sequence for each scene, while the training split is constructed by taking either the middle segment of the GT sequence or a temporally sampled version of that middle segment.

SeeU45 covers a diverse set of conditions in terms of scene type, foreground subjects, camera regimes, and motion types. A summary of the dataset statistics is provided in Table~\ref{tab:seeu45_stats}.
\setcounter{table}{3}

\begin{table}[t]
    \centering
    \small
    \setlength{\tabcolsep}{6pt} 
    \caption{Statistics of the SeeU45 dataset.}
    \label{tab:seeu45_stats}
    \begin{tabular}{@{}l l@{}}
        \toprule
        \multicolumn{2}{c}{\textbf{Composition}} \\
        \midrule
        Scenes (total)                  & 45 \\
        Our captured               & 10 \\
        From public datasets            & 35 \\
        \midrule
        \multicolumn{2}{c}{\textbf{Scene Types}} \\
        \midrule
        Indoor scenes                     & 6 \\
        Outdoor scenes                    & 39 \\
        \midrule
        \multicolumn{2}{c}{\textbf{Foreground Subjects (some scenes contain both)}} \\
        \midrule
        Humans                            & 8 \\
        Animals                           & 13 \\
        Robots                            & 3 \\
        Vehicles                          & 18 \\
        Everyday objects                  & 12 \\
        \midrule
        \multicolumn{2}{c}{\textbf{Camera Regimes}} \\
        \midrule
        Static                            & 10 \\
        Handheld                          & 28 \\
        Drone                             & 7 \\
        \midrule
        \multicolumn{2}{c}{\textbf{Motion Types}} \\
        \midrule
        Rigid motion                      & 21 \\
        Non-rigid motion                  & 24 \\
        \midrule
        \multicolumn{2}{c}{\textbf{Frame Statistics}} \\
        \midrule
        GT frames / scene (min / max / avg) 
                                          & 9 / 521 / 80.24 \\
        Train frames / scene (min / max / avg) 
                                          & 7 / 47 / 15.96 \\
        \bottomrule
    \end{tabular}
\end{table}


\section{More Details of Continuous 4D Dynamics Model}
\label{sec:supp_C4DD}
\subsection{Architecture.} In Algorithm \ref{alg:C4DD} we present the detailed architecture of the Continuous 4D Dynamics Model (C4DD). During training, C4DD learns continuous and smooth motion/camera bases by optimizing B-spline control points, and enforces physically consistent and smooth extrapolation through physics-aware constraints.

\begin{algorithm}[t]
\caption{Continuous 4D Dynamics Model Architecture}
\label{alg:C4DD}
\begin{algorithmic}[1]
\Require
    Number of motion bases $K$, number of control points $M$, degree $p$.
\Statex

\State \textbf{Initialization:}
\State Build an open-uniform knot vector $\{u_i\}_{i=0}^{M+p}$ on $[0,1]$.
\State Initialize motion control points 
       $\mathbf{C}^{\text{mot}} \in \mathbb{R}^{K \times 9 \times M}$.
\State Initialize camera control points 
       $\mathbf{C}^{\text{cam}} \in \mathbb{R}^{1 \times 9 \times M}$.

\Statex
\Function{BSplineBasis}{$\mathbf{t}_{01}$}
    \Comment{$\mathbf{t}_{01} \in (0,1)^{B\times 1}$}
    \State Compute B-spline basis $\mathbf{B} \in \mathbb{R}^{B \times M}$ 
           using Cox--de Boor recursion on knots $\{u_i\}$.
    \State \Return $\mathbf{B}$
\EndFunction

\Statex
\Function{Forward}{$\mathbf{t}_{\mathrm{norm}}$}
    \Comment{$\mathbf{t}_{\mathrm{norm}} \in [-1,1]^{B\times 1}$}
    \State Map time to $(0,1)$:
           $\mathbf{t}_{01} \gets \mathrm{clip}(0.5\,\mathbf{t}_{\mathrm{norm}} + 0.5,\,10^{-6},\,1-10^{-6})$.
    \State $\mathbf{B} \gets \Call{BSplineBasis}{\mathbf{t}_{01}}$ \Comment{$\mathbf{B} \in \mathbb{R}^{B\times M}$}
    \State \textbf{Motion bases:}
           $\mathbf{Y}^{\text{mot}} \gets \mathbf{C}^{\text{mot}} \mathbf{B}^\top \in \mathbb{R}^{K\times 9\times B}$.
    \State \textbf{Camera pose:}
           $\mathbf{Y}^{\text{cam}} \gets \mathbf{C}^{\text{cam}} \mathbf{B}^\top \in \mathbb{R}^{1\times 9\times B}$.
    \State Reshape to $\mathbf{Y}^{\text{mot}} \in \mathbb{R}^{K\times B\times 9}$,
           $\mathbf{Y}^{\text{cam}} \in \mathbb{R}^{B\times 9}$.
    \State \Return $\mathbf{Y}^{\text{mot}},\; \mathbf{Y}^{\text{cam}}$
\EndFunction

\Statex
\Function{ForwardExtrap}{$\mathbf{t}_{\mathrm{norm}}$}
    \Comment{linear extrapolation outside $[-1,1]$}
    \State $\mathbf{t}_{\mathrm{clamp}} \gets \mathrm{clip}(\mathbf{t}_{\mathrm{norm}}, -1, 1)$
    \State $\mathbf{Y}_0^{\text{mot}}, \mathbf{Y}_0^{\text{cam}} 
           \gets \Call{Forward}{\mathbf{t}_{\mathrm{clamp}}}$
    \State Estimate endpoint slopes $\mathbf{s}_L, \mathbf{s}_R$ 
           at $t=-1$ and $t=1$ using finite differences 
           (step size $\Delta t \approx 2/(T-1)$).
    \State Initialize 
           $\mathbf{Y}^{\text{mot}} \gets \mathbf{Y}_0^{\text{mot}}$,
           $\mathbf{Y}^{\text{cam}} \gets \mathbf{Y}_0^{\text{cam}}$.
    \For{each time index $b$}
        \If{$t_{\mathrm{norm}}[b] < -1$}
            \State $\Delta t \gets t_{\mathrm{norm}}[b] + 1$
            \State Extrapolate 
                   $\mathbf{Y}^{\text{mot}}[:,b,:]$ and $\mathbf{Y}^{\text{cam}}[b,:]$ 
                   using left endpoint and slope $\mathbf{s}_L$.
        \ElsIf{$t_{\mathrm{norm}}[b] > 1$}
            \State $\Delta t \gets t_{\mathrm{norm}}[b] - 1$
            \State Extrapolate 
                   $\mathbf{Y}^{\text{mot}}[:,b,:]$ and $\mathbf{Y}^{\text{cam}}[b,:]$ 
                   using right endpoint and slope $\mathbf{s}_R$.
        \EndIf
    \EndFor
    \State \Return $\mathbf{Y}^{\text{mot}},\; \mathbf{Y}^{\text{cam}}$
\EndFunction

\end{algorithmic}
\end{algorithm}

\subsection{More Details About Ablation Study on C4DD Architecture.} 
\begin{figure*}[h]
\centering
\includegraphics[width=1.0\linewidth, trim={0 0 0 0}, clip]{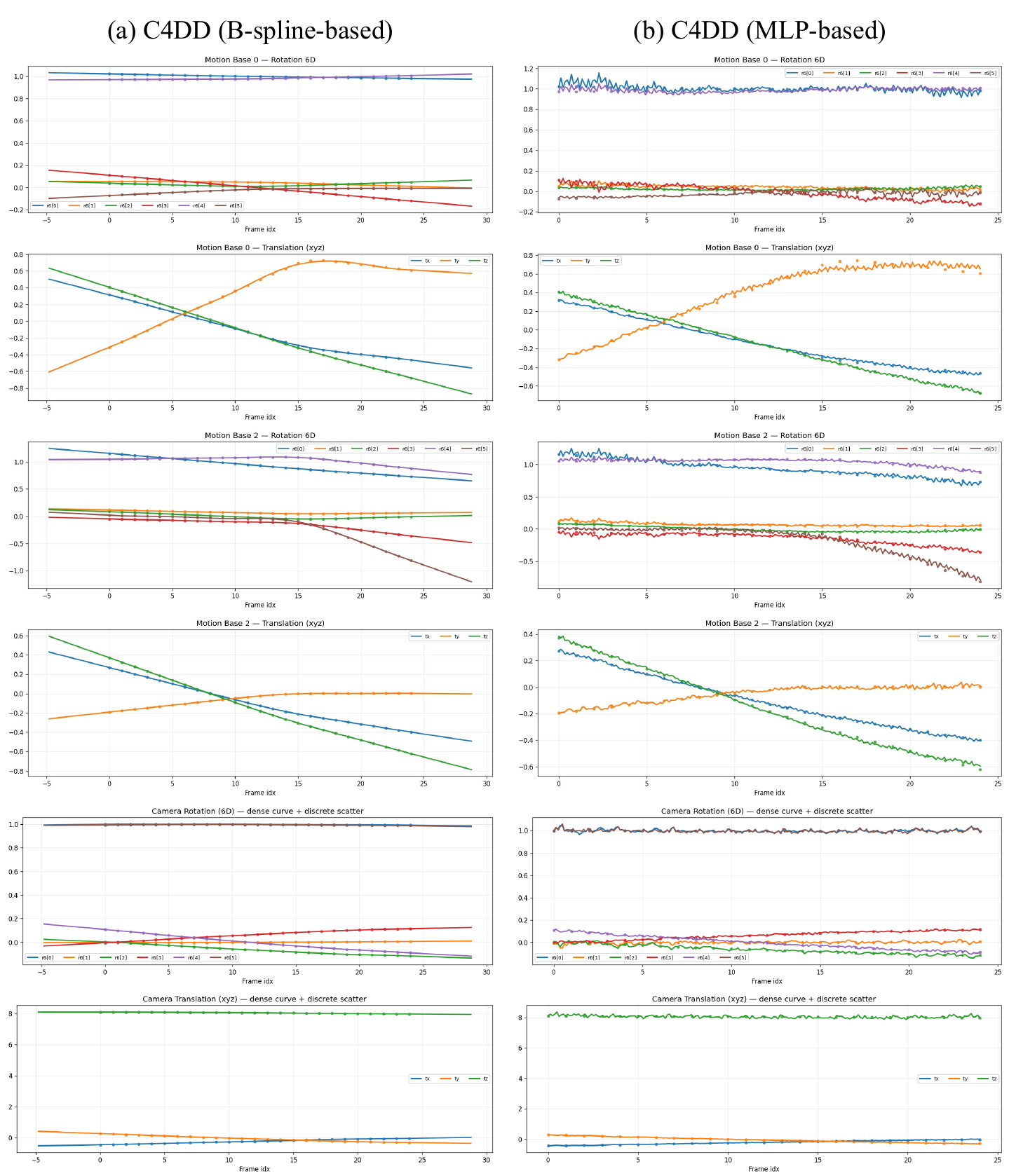}
\caption{The C4DD with MLP variant (b) predicts motion bases with reduced temporal smoothness and physical consistency, confirming the advantageous inductive bias of spline priors (a) for continuous dynamics.}
\label{fig:mlp_comp}
\end{figure*}

\begin{figure}[h]
\centering
\includegraphics[width=1.0\linewidth, trim={0 0 0 0}, clip]{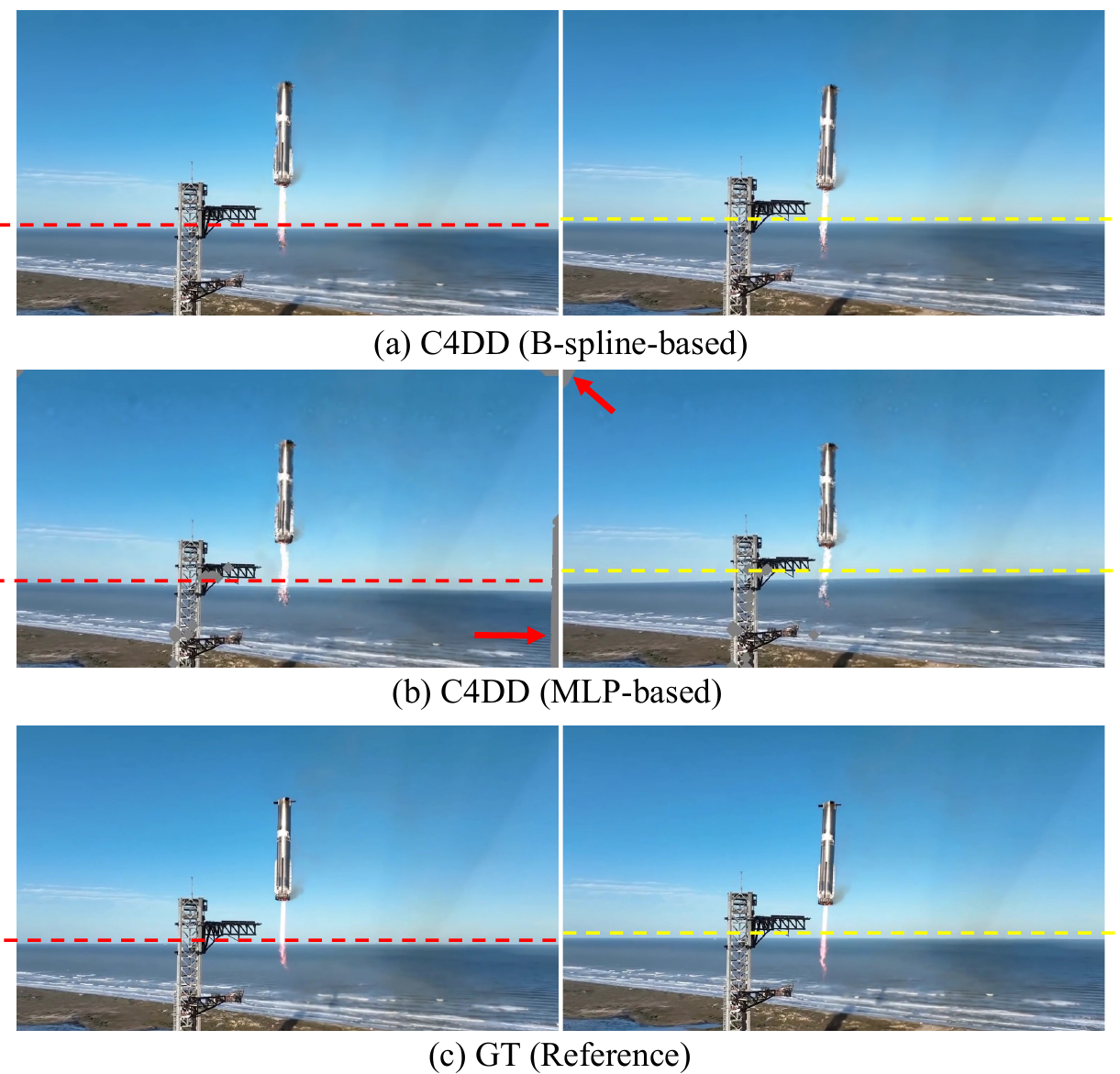}
\caption{Visual comparison on C4DD Architectures. The C4DD with spline constrains (a) has better smoothness and physical consistency (both camera pose and foreground dynamics).}
\label{fig:mlp_visual}
\end{figure}

To assess the necessity of the proposed C4DD architecture, we replace the B-spline–based design with a pure MLP layer and train it thoroughly. As shown in Fig. \ref{fig:mlp_comp}, although the MLP-based variant can roughly fit the overall trend of the motion bases, the resulting continuous trajectories are highly noisy and lack smoothness. This significantly degrades the spatio-temporal quality of the model’s predictions (see Fig. \ref{fig:mlp_visual}).

\section{More Details of Spatial-Temporal In-Context Generation}
\label{sec:supp_context}
The third stage, context-aware video filling, is mainly adapted from VACE \cite{Jiang_2025_VACE}. During inference, We provide three types of contextual information:

\begin{enumerate}
    \item \textbf{Text prompt.} The text prompt describes the global spatio-temporal semantics, adds extra guidance for the regions to be filled, and emphasizes physical consistency in the scene. For example: 
    \emph{``A camel is walking slowly in his enclosure. The enclosure has sand on the floor, surrounded by a wooden fence and planks. The background has trees. Restore the masked regions of the video with the background of the enclosure. Make the colors and background behind the camel realistic and continuous.''}

    \item \textbf{Projected frames.} These frames come from our $4\text{D}\rightarrow 2\text{D}$ rendering. Pixels in the inpainting masks (defined below) are set to a constant gray value (127), so that the projected frames act as a structural scaffold for the video while clearly indicating where content to be synthesized.

    \item \textbf{Inpainting masks.} The masks specify the unseen regions that need to be filled. They are constructed from three types of areas: (1) regions that are never observed (novel viewpoints or previously occluded areas), which are naturally identified through the inverse-projection process; (2) locations where the projected Gaussians have low confidence, detected via a threshold on the opacity values; and (3) thin structures and sharp depth discontinuities that may cause projection artifacts (e.g., along object boundaries and occlusion edges), detected by checking whether the relative depth difference exceeds a predefined threshold.
\end{enumerate}

\section{More Details of Proposed Metrics}
\label{sec:supp_metrics}

To evaluate the spatial consistency of generated videos in unseen viewpoints, 
we adopt two standard two-view geometric metrics: 
\emph{Epipolar Error (EE)} and \emph{Epipolar Inlier Ratio (EIR)}.  
These metrics quantify how well the generated frames obey the underlying 
epipolar geometry defined by a fundamental matrix estimated from visual correspondences.

\textbf{Setup.} Given a reference frame $I_1$ and a generated frame $I_2$, 
we extract putative feature correspondences 
$\{(x_1^{(i)}, x_2^{(i)})\}_{i=1}^N$ using SIFT with cross-check and ratio test.
The fundamental matrix $F$ is then estimated via RANSAC:
\begin{equation}
    F = \arg\min_{F'} 
    \sum_{i \in \mathcal{I}(F')} 
    \mathrm{EE}(x_1^{(i)}, x_2^{(i)}, F'),
\end{equation}
where $\mathcal{I}(F')$ denotes the RANSAC inlier set.

\textbf{Epipolar Error (EE).}
For a correspondence $(x_1, x_2)$ with homogeneous coordinates 
$\tilde{x}_1 = (x_1^\top, 1)^\top$ and $\tilde{x}_2 = (x_2^\top, 1)^\top$, 
the epipolar constraint states:
\begin{equation}
    \tilde{x}_2^\top F \tilde{x}_1 = 0.
\end{equation}
Deviations from this constraint reflect geometric inconsistency.
We adopt the \emph{Sampson approximation} of the re-projection error:
\begin{equation}
\mathrm{EE}(x_1, x_2; F)
=
\sqrt{
    \frac{
        (\tilde{x}_2^\top F \tilde{x}_1)^2
    }{
        (F\tilde{x}_1)_0^2 
        + (F\tilde{x}_1)_1^2 
        + (F^\top \tilde{x}_2)_0^2 
        + (F^\top \tilde{x}_2)_1^2
    }
}.
\end{equation}
This metric has several desirable properties:
\begin{itemize}
    \item It is expressed in pixel units and directly interpretable.
    \item It approximates the \emph{geometric reprojection error} without requiring camera intrinsics.
    \item It is robust to scale ambiguity inherent to fundamental matrices.
\end{itemize}
In practice, we report:
\[
\mathrm{EE}_{\mathrm{median}} = 
\mathrm{median}_{i \in \mathcal{I}(F)}\;\mathrm{EE}(x_1^{(i)}, x_2^{(i)}; F),
\]
where a lower value indicates better geometric alignment between the generated view and the reference view.

\textbf{Epipolar Inlier Ratio (EIR).}
While EE measures the \emph{accuracy} of the geometric alignment, 
we also evaluate the \emph{stability} of the geometry via an inlier ratio:
\begin{equation}
    \mathrm{EIR}
    = 
    \frac{
        |\mathcal{I}(F)|
    }{
        N
    },
\end{equation}
where $N$ is the total number of matched correspondences before RANSAC.
A higher EIR indicates that a larger portion of correspondences can be explained by a 
valid two-view geometry, suggesting more realistic spatial structure in the generated frame.

EIR is particularly informative in generative settings because:
\begin{itemize}
    \item generated videos may contain distortions that destroy epipolar geometry,
    \item RANSAC tends to reject correspondences that violate 3D plausibility,
    \item a low EIR implies strong geometric hallucination or temporal drift.
\end{itemize}

Together, EE and EIR evaluate the spatial fidelity of generated videos:
\begin{itemize}
    \item \textbf{EE} reflects how close the video is to a physically plausible two-view geometry.
    \item \textbf{EIR} reflects how consistently the generator maintains global 3D structure.
\end{itemize}
Both metrics do not require camera intrinsics and thus can be applied to arbitrary videos, 
making them suitable for evaluating geometry realism in this work.

\section{Limitations}
\label{sec:supp_limitations}

\begin{figure}[h]
\centering
\includegraphics[width=1.0\linewidth, trim={0 0 0 0}, clip]{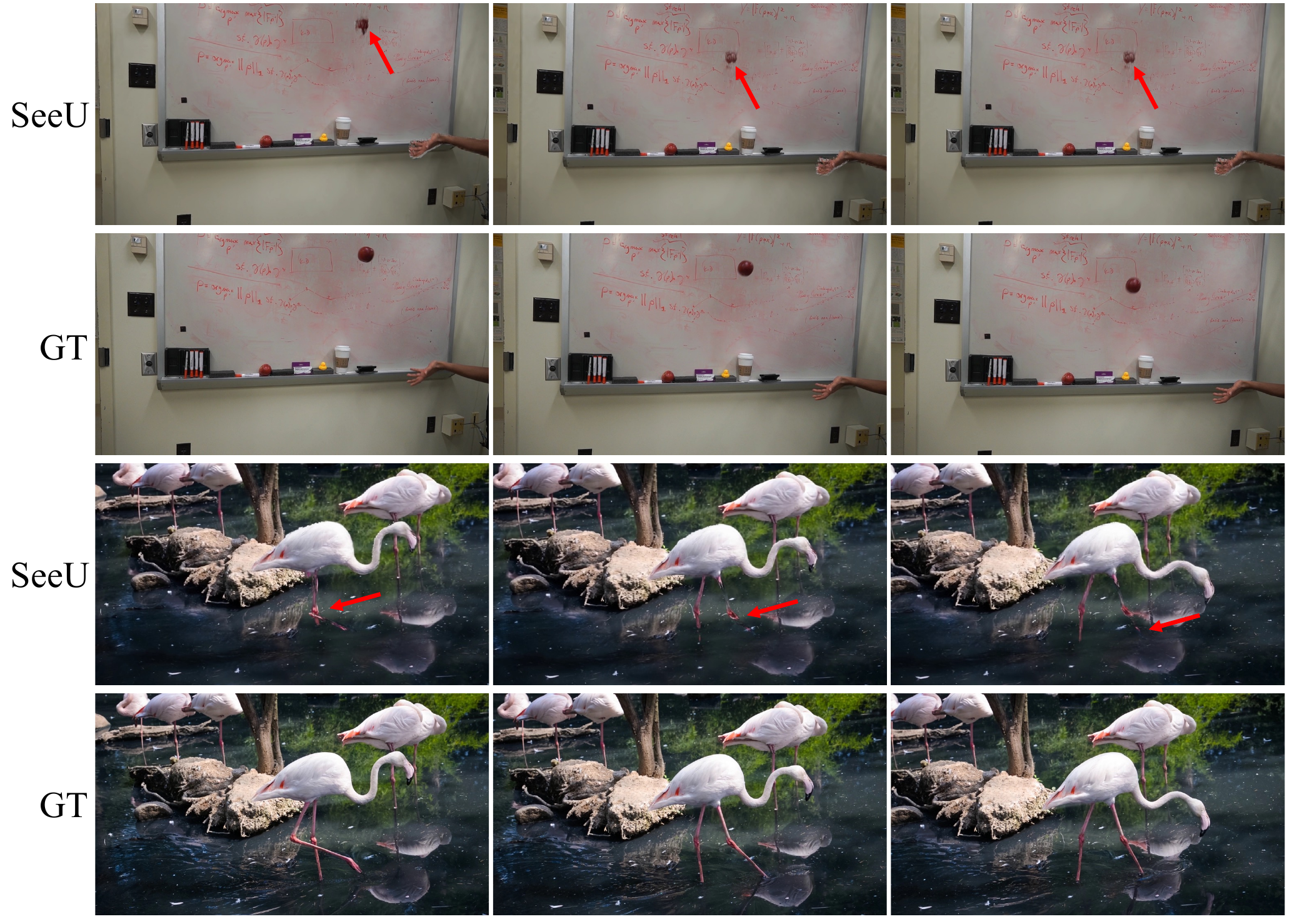}
\caption{SeeU’s performance degrades on inputs containing thin structures or lacking texture, reflecting the inherent limitations of existing base models.}
\label{fig:limit}
\end{figure}

In the first-stage 2D-4D lifting, the performance of SeeU is strongly constrained by the quality of the upstream geometry modules, including camera pose estimation, tracking, and depth prediction. As a result, SeeU requires input videos with salient foreground objects and sufficient spatial details.

As shown in Fig. \ref{fig:limit}, when the foreground is extremely small or lacks rich texture, these modules become unreliable and the final outputs degrade accordingly. We illustrate such failure cases on small or low-texture foregrounds in the examples below.

\section{Robustness and Scalability Analysis}
\label{sec:supp_complex}

\begin{figure*}[h]
\centering
\includegraphics[width=\linewidth]{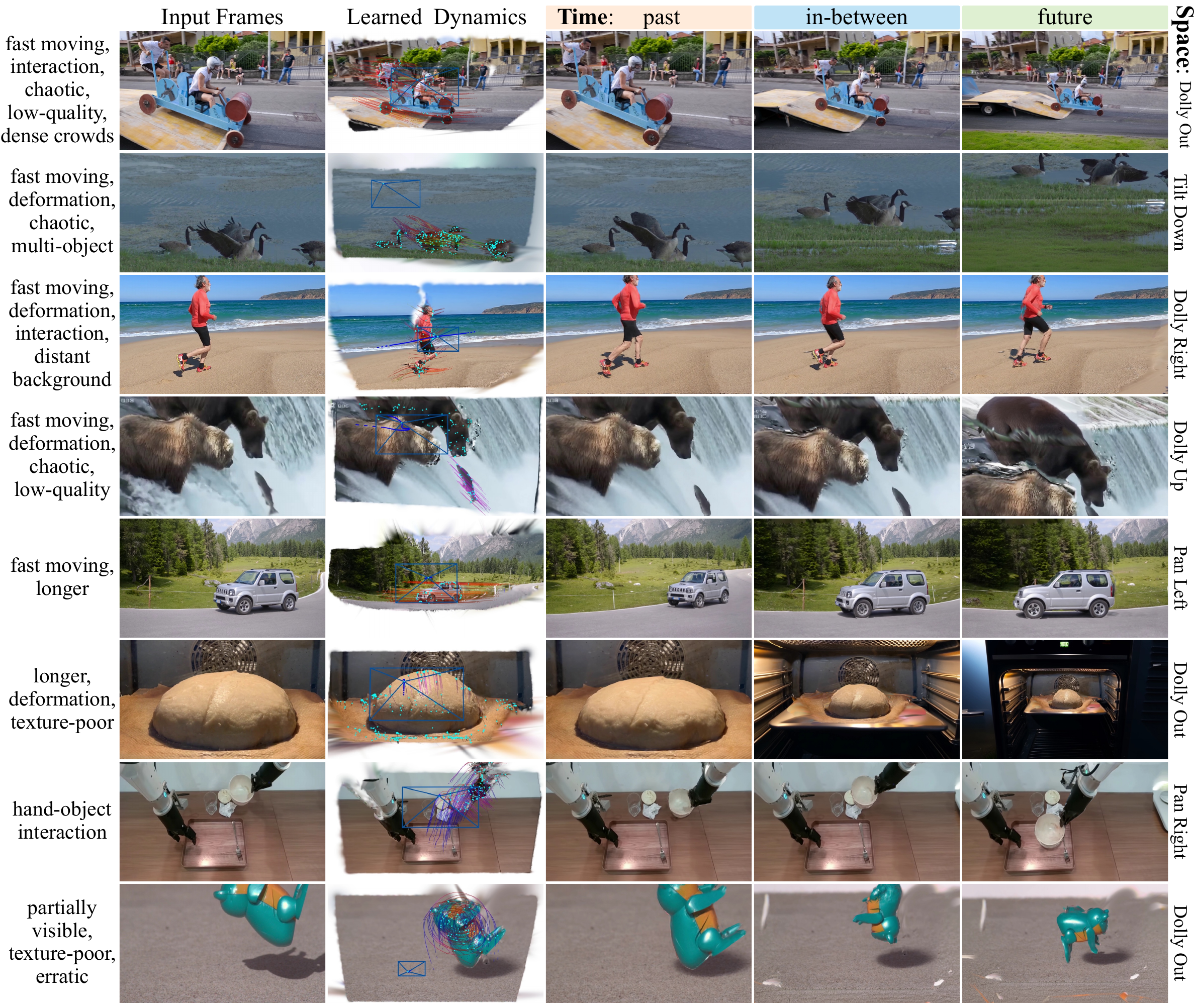}
\caption{\small{Robustness evaluations (please zoom in).}}
\label{fig: complex}
\end{figure*}

In Fig. \ref{fig: complex}, we show additional experiments under more challenging cases. These extreme cases involve combinations of adverse factors, including fast-moving cameras or objects, interactions, chaotic dynamics, low-quality inputs, dense crowds, deformations, multiple objects, longer time span, distant backgrounds, and texture-poor scenes (first column). The results show that SeeU can \textbf{effectively learn continuous 4D dynamics (third column)} and generate coherent outputs in unseen time and space, although some artifacts may remain. Additionally, we believe that longer and more complex videos can be decomposed into temporally stable chunks and processed sequentially by SeeU.

\section{More Comparison Results}
\label{sec:supp_comparison}

We compare our method with two methods that follow similar 3D/4D memory-based designs, namely DaS \cite{Gu_2025_Das} and HunyuanWorld-Voyager \cite{Huang_2025_Voyager}. As shown in Table \ref{table:number}, our method achieves better performance in both geometric consistency and semantic quality. While all these models inject geometric priors, SeeU \textbf{additionally models temporal dynamics}. We will discuss and compare with these approaches in the revised manuscript.

\begin{table}[t]
  \caption{Comparison between 4D-aware models.}
  \label{table:number}
  \scriptsize
  \centering
  \begin{tabularx}{\columnwidth}{
    l
    >{\centering\arraybackslash}p{0.16\columnwidth}
    >{\centering\arraybackslash}p{0.16\columnwidth}
    >{\centering\arraybackslash}p{0.3\columnwidth}
  }
    \toprule
    Method & SeeU & DaS & HunyuanWorld-Voyager  \\
    \midrule
    EIR$\uparrow$/CLIP-V$\uparrow$ & \textbf{0.8024}/\textbf{0.9588} & 0.7624/0.9253 & 0.7298/0.8911 \\
    \bottomrule
  \end{tabularx}
\end{table}

\section{More Visual Results}
\label{sec:supp_results}
\begin{figure*}[h]
\centering
\includegraphics[width=1.0\linewidth, trim={0 0 0 0}, clip]{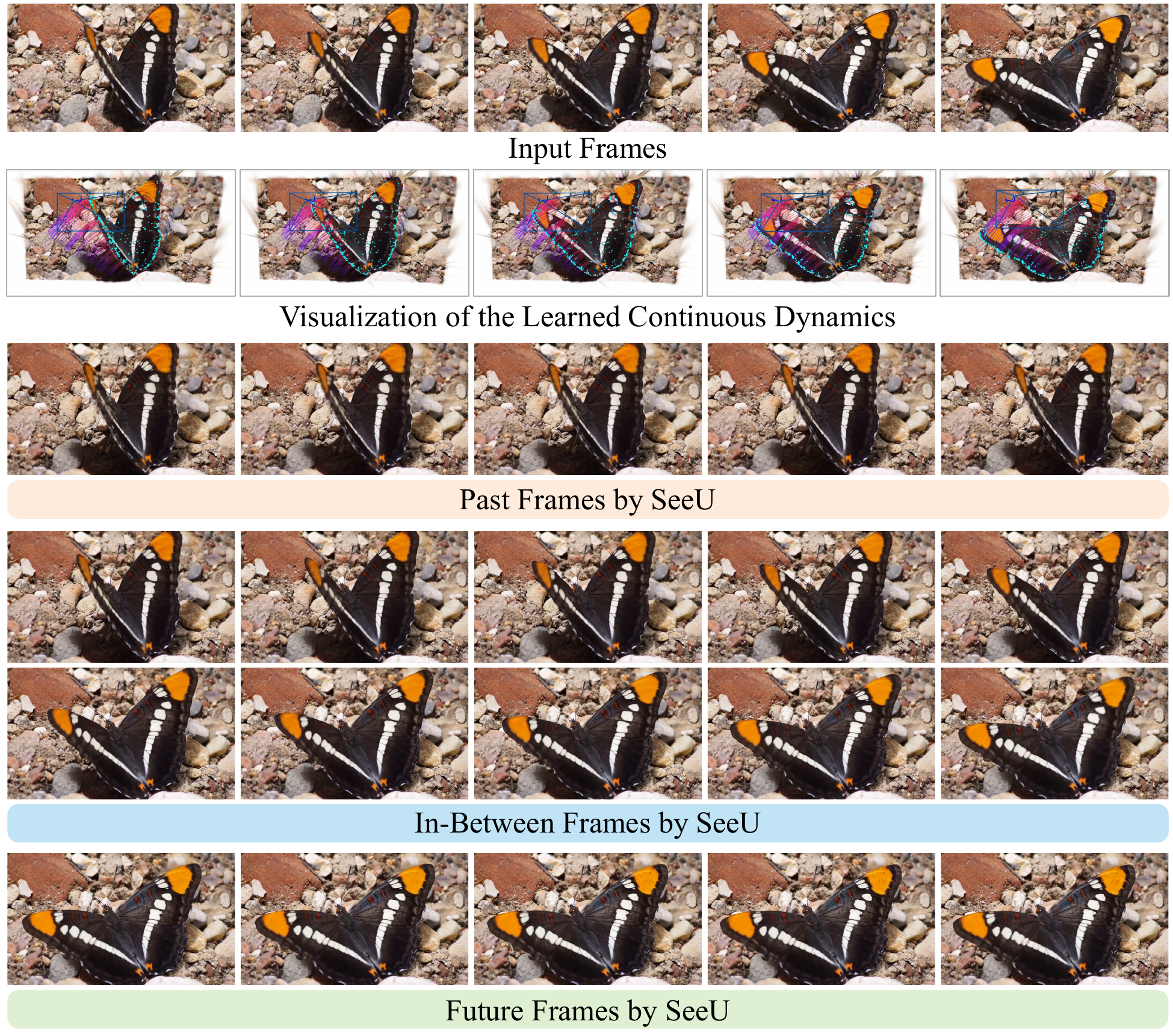}
\caption{Unseen Temporal World Generated by SeeU.}
\label{fig:supp_time}
\end{figure*}

\begin{figure*}[h]
\centering
\includegraphics[width=1.0\linewidth, trim={0 0 0 0}, clip]{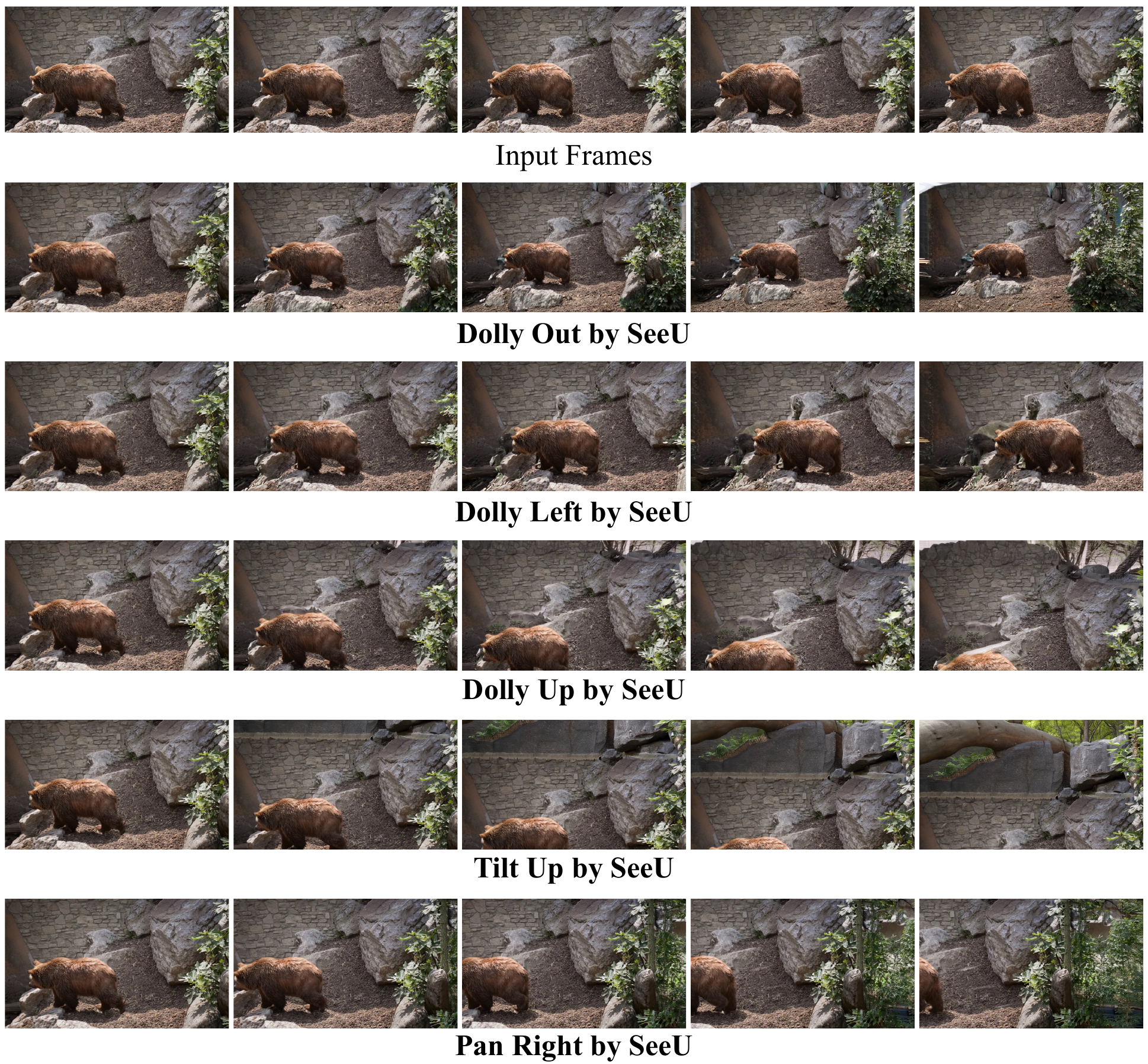}
\caption{Unseen Spatial World Generated by SeeU.}
\label{fig:supp_space}
\end{figure*}

\begin{figure*}[h]
\centering
\includegraphics[width=1.0\linewidth, trim={0 0 0 0}, clip]{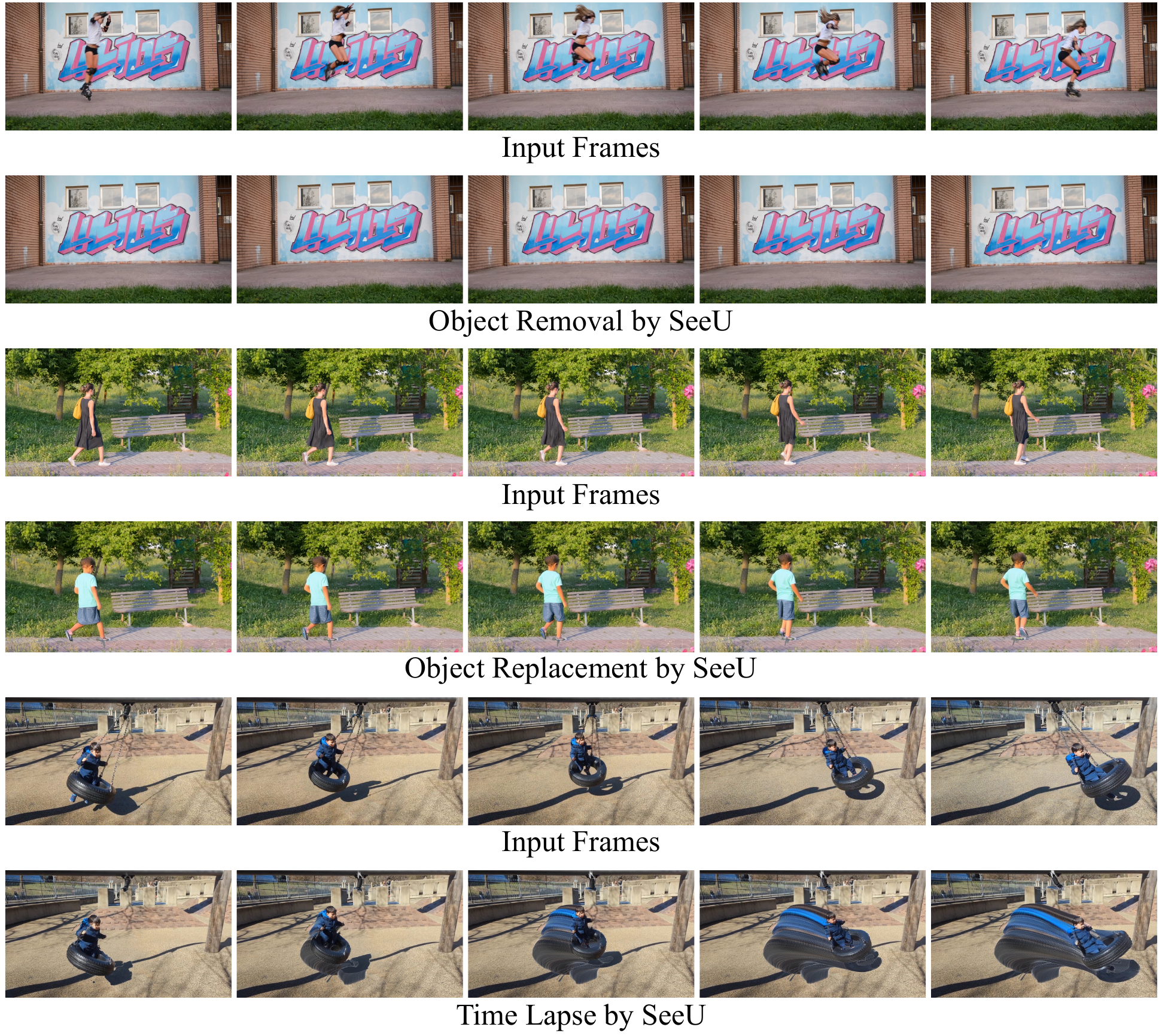}
\caption{Videos edited by SeeU.}
\label{fig:supp_edit}
\end{figure*}

We encourage readers to directly watch the \textbf{videos} provided in the supplementary materials, as they best demonstrate the temporal and spatial behaviors of our method. For completeness, we also include a few representative visual examples below (Fig. \ref{fig:supp_time} to Fig. \ref{fig:supp_edit}) as a preview.

\end{document}